%% file: arxiv_main.tex
\documentclass[wcp]{jmlr}

\usepackage{booktabs}
\usepackage{colortbl}  % \columncolor for shading Table 2 split-groups
\usepackage{xspace}
\usepackage{placeins}
\usepackage{float}
\usepackage{multirow}
\usepackage{wrapfig}

\usepackage{lineno}
\makeatletter
\let\Ginclude@graphics\@org@Ginclude@graphics
\makeatother

\newcommand{\figtightcap}{\vspace{-8pt}}

\jmlrvolume{353,}
\jmlryear{2026}
\jmlrworkshop{ACML 2026}

\title[MINER]{MINER: Multi-crop INference-time Enhancement for Rare-Object Retrieval with Frozen Dual Encoders}

\author{%
  \Name{Abdulmalik Alquwayfili\nametag{\thanks{Corresponding author}}} \Email{aalquwayfili@ncai.gov.sa}\\
  \Name{Faisal AlMeshal} \Email{falmeshal@ncai.gov.sa}\\
  \Name{Jumanah Almajnouni} \Email{jalmajnouni@ncai.gov.sa}\\
  \Name{Huda Abdulhadi Alamri} \Email{haamri@ncai.gov.sa}\\
  \Name{Muhammad Kamran J Khan} \Email{mkkhan@ncai.gov.sa}\\
  \addr Saudi Data and Artificial Intelligence Authority (SDAIA)}

\editors{Andy Song, Bo Han and Sarah Erfani}

\begin{document}

\makeatletter
\let \@jmlrpages \@empty
\makeatother

\maketitle

\begin{abstract}
Text-to-image retrieval with frozen dual encoders degrades when the query names a small, visually subordinate object in a cluttered scene: a single global image embedding underrepresents the localized visual evidence. We present \textbf{MINER}, a training-free inference framework that augments a frozen dual encoder's global image embedding with a small bank of region-level embeddings and a hubness-correcting similarity rescoring, recovering visual evidence that global pooling underweights. To evaluate this setting, we introduce \textbf{ROCS}, a benchmark built from high-clutter subsets of Flickr30K and MS COCO whose images are re-captioned to name a single low-salience object. Experiments on CLIP, SigLIP, and SigLIP\,2 show that MINER improves retrieval on every backbone, on ROCS and on the standard splits. Analyses show that these gains come primarily from broader spatial coverage rather than precise crop placement, revealing a simple and general way to recover localized evidence from frozen representations.
Code: \url{https://github.com/aalquwayfili/MINER}. Dataset: \url{https://huggingface.co/datasets/aalquwayfili/ROCS}.
\end{abstract}

\begin{keywords}
text-to-image retrieval; vision-language models; dual encoders; training-free inference; region augmentation; hubness correction
\end{keywords}

\input{sec/1_intro}
\input{sec/2_related_work}
\FloatBarrier
\input{sec/3_dataset}

\input{sec/4_method}
\input{sec/5_experiments}
\FloatBarrier
\input{sec/6_conclusion}
\FloatBarrier

%\acks{Acknowledgements should go here in the camera-ready version.}

% ACML counts references toward the 16-page limit, so tighten inter-entry
% spacing rather than dropping citations.
\clearpage
{\footnotesize \setlength{\bibsep}{0pt}\linespread{0.97}\selectfont\bibliography{egbib}}

\clearpage
\appendix
\input{sec/appendix_algorithm}
\input{sec/appendix_sweeps}
\input{sec/appendix_pooling_probe}
\input{sec/appendix_detailed}
\input{sec/appendix_audit}
\input{sec/appendix_figures}
\input{sec/appendix_prompts}
\input{sec/appendix_cropfig}

\end{document}

%% file: sec/1_intro.tex
\section{Introduction}

\begin{figure}[t]
\centering
% 12% wider than the text block, centred: 0.36in into each 1.25in margin.
% \makebox declares one text width so no Overfull warning is raised.
\makebox[\textwidth][c]{%
  \includegraphics[width=1.12\textwidth]{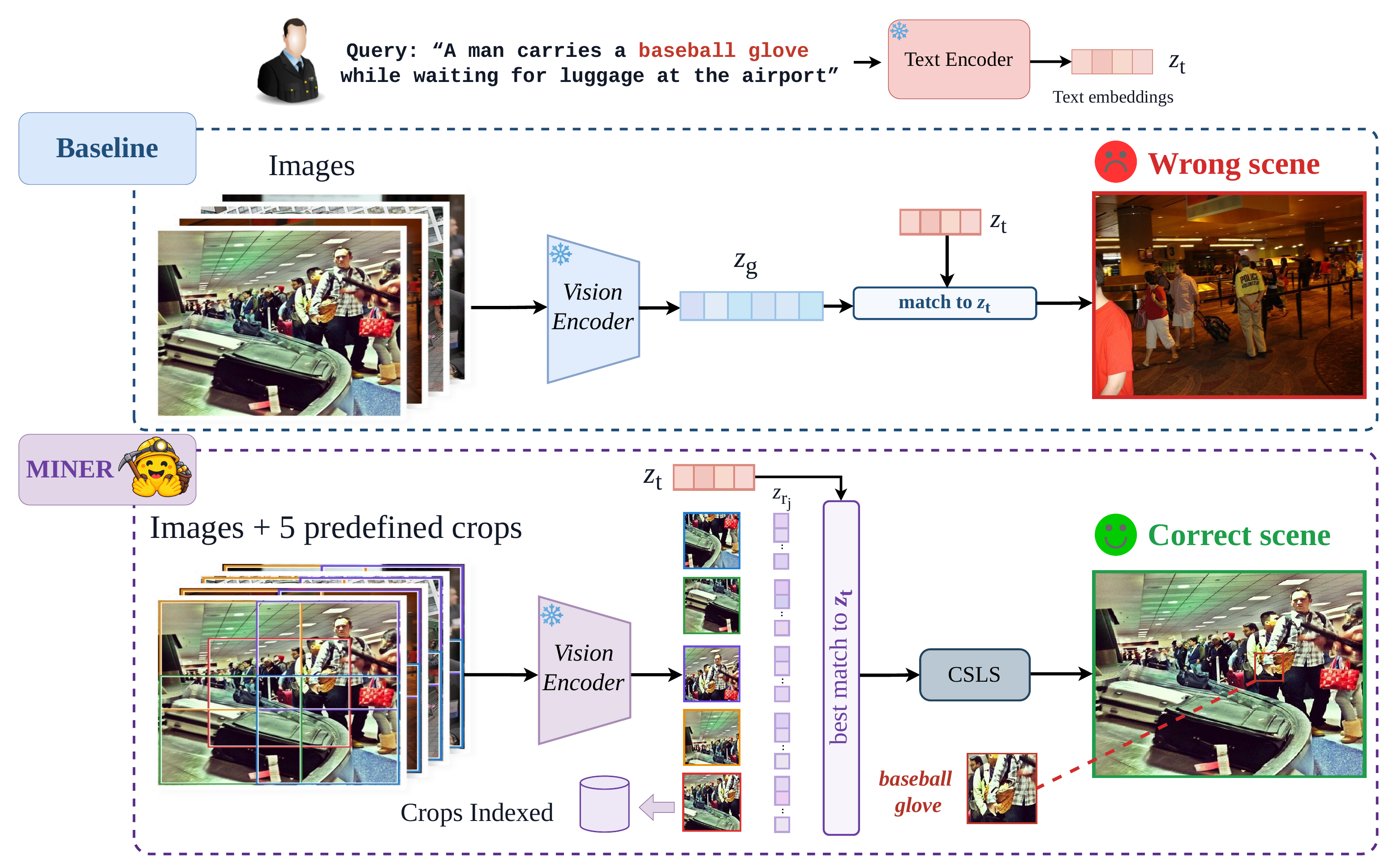}}
\caption{\textbf{One global vector per image is not enough.} Both panels
share the same frozen vision encoder. The baseline scores each gallery
image by its global embedding alone and retrieves a similar but incorrect
scene; MINER adds five predefined crops per image, blends the best crop
score with the global one, and rescores with two-sided CSLS. The queried
baseball glove covers under $2\%$ of the frame.}
\label{fig:teaser}
\end{figure}

Text-to-image retrieval, the task of ranking images in a large
collection by their match to a natural-language query, is a core
building block of multimedia search, content recommendation, and
large-scale visual indexing~\citep{radford2021learning,jia2021scaling,zhai2023siglip}. Recent
advances in vision--language pretraining have made dual-encoder models
the dominant paradigm: CLIP~\citep{radford2021learning},
SigLIP~\citep{zhai2023siglip}, and SigLIP\,2~\citep{tschannen2025siglip}
project each modality independently into a shared embedding space, and
retrieval is performed by cosine similarity.

The dual-encoder architecture is efficient but compresses every image
into a single global vector. In densely populated scenes, the correct
match may depend on small or visually peripheral objects whose
localized evidence is poorly captured by a single global
embedding~\citep{wang2023sclip,yao2021filip}. Captions that single out
such objects therefore expose a failure mode of global pooling that
standard retrieval benchmarks do not stress (Figure~\ref{fig:teaser}).

To address this failure mode we present \textbf{MINER} (Multi-crop
INference-time Enhancement for Rare-object Retrieval), a training-free inference
pipeline for frozen dual encoders. MINER re-encodes a small bank of fixed
image crops through the same frozen backbone, blends the strongest
regional similarity with the global one, and refines the resulting
similarity matrix with a retrieval-space rescoring step that corrects
hubness. The pipeline is training-free: our analysis shows that what region
augmentation contributes is governed by spatial coverage, not
localization, so a parameter-free fixed crop set matches every
attention-guided alternative we tested.

Prior fine-grained retrieval approaches often rely on retraining
region-aware representations or token-level
alignment~\citep{yao2021filip,zhong2021regionclip}, while other recent
methods require query-conditioned inference or per-query
attribution~\citep{zhan2025elip,zhao2025gradeclip}. These assumptions
are poorly aligned with frozen dual-encoder retrieval, where only
pretrained embeddings are available at inference time and each gallery
image must be encoded once and reused across many queries, making
query-conditioned region extraction impractical at scale. MINER instead
requires only the frozen encoder: every gallery image and its crops are
encoded once at indexing time and reused across all queries.

Our main contributions are summarized as follows:
\begin{itemize}

\item \textbf{MINER: A lightweight training-free retrieval framework.}
We propose \textbf{MINER}, an inference-time framework that enhances frozen vision--language retrieval models by combining global image representations with complementary region-level embeddings and two-sided CSLS re-scoring. MINER requires no retraining, query conditioning, or auxiliary models at inference, making it readily applicable to existing dual-encoder architectures.

\item \textbf{ROCS: A benchmark for retrieval in cluttered scenes.}
We introduce \textbf{ROCS} (Rare Objects in Cluttered Scenes), a benchmark constructed from high-clutter subsets of Flickr30K and MS COCO. Images are curated to contain crowded scenes and re-captioned to emphasize rare, low-attention, and visually subordinate objects, providing a challenging evaluation setting for image--text retrieval.

\item \textbf{Design principles for training-free retrieval.}
Through extensive experiments on CLIP, SigLIP, and SigLIP\,2, we demonstrate that MINER consistently improves retrieval performance across standard benchmarks and ROCS. Our analysis further shows that recovering small-object recall is driven by spatial coverage rather than by precise crop localization, which is what makes a parameter-free fixed-crop strategy sufficient: it matches attention-guided alternatives while preserving strong generalization across backbone architectures.

\end{itemize}

%% file: sec/2_related_work.tex
\section{Related Work}
\label{sec:related-work}

\subsection{Text-to-Image Retrieval}

Large-scale vision--language pretraining on web image--text pairs drives current text-to-image retrieval~\citep{radford2021learning,jia2021scaling,li2022blip,zhai2023siglip,zhan2025elip}.

Dual-encoder architectures have become the dominant paradigm for this task. Models such as CLIP and ALIGN learn aligned image and text representations using contrastive learning over large-scale datasets, enabling strong zero-shot retrieval performance across multiple benchmarks~\citep{radford2021learning,jia2021scaling}. Subsequent works have further improved representation quality and training efficiency: BLIP introduces bootstrapped caption generation~\citep{li2022blip}, and SigLIP replaces the softmax contrastive loss with a sigmoid loss to improve scalability~\citep{zhai2023siglip}. These models compress each image into a single global embedding, which can underrepresent small or visually subordinate objects.

\subsection{Fine-Grained Vision--Language Alignment}

To address the limitations of global representations, several works explore fine-grained alignment between image regions and textual tokens. FILIP introduces a late-interaction mechanism that computes token-level similarity between image patches and textual tokens, enabling finer-grained cross-modal alignment while maintaining efficient inference~\citep{yao2021filip}. PyramidCLIP aligns hierarchical features at several granularities~\citep{gao2022pyramidclip}.

Another line of work focuses on region-level representations. RegionCLIP extends contrastive language-image pretraining to region-based representations, enabling alignment between textual concepts and localized image regions~\citep{zhong2021regionclip}. More recently, methods such as ELIP introduce lightweight text-guided visual prompts that condition the image encoder on the query, improving retrieval performance without retraining large backbone models~\citep{zhan2025elip}.

A further alternative is to re-rank a dual encoder's shortlist with a stronger trained model: cross-attention re-rankers recover accuracy at the cost of a per-query forward pass over every shortlisted candidate~\citep{miech2021thinking}, and query-conditioned encoders such as ELIP, above, must re-encode the top-ranked images for every query. MINER makes the opposite trade: it is training-free and query-agnostic, so every gallery image is encoded once at indexing time and reused across all queries.

\subsection{Retrieval Benchmarks}

Closest to our benchmark are efforts that renovate the standard splits themselves: MSCOCO-FG and Flickr30K-FG rewrite coarse captions into fine-grained descriptions and renovate the candidate image pools~\citep{chen2023rethinking}, and a recent reproducibility study shows that retrieval models are sensitive to exactly this caption granularity~\citep{hendriksen2025benchmark}. These benchmarks refine the full scene description, whereas ROCS (Section~\ref{sec:rocs}) isolates a different failure mode: segmentation-based filtering selects cluttered scenes, and each caption is re-anchored on a single small, rare object. The two directions are complementary: FG-style renovation stresses how precisely a caption is matched, while ROCS stresses whether a low-salience object is represented at all.

\subsection{Multi-Crop Inference Strategy}

Test-time augmentation by multi-crop is a long-standing technique in visual recognition. \citet{krizhevsky2012alexnet} introduced the five-crop strategy (four corners and a center), \citet{szegedy2015googlenet} extended it to dense multi-scale grids of up to 144 crops, and \citet{he2016resnet} standardised multi-crop evaluation for deep networks. These crops are query-agnostic and geometrically fixed: they are placed without regard to image content or the query.

More recent work makes cropping content-aware. \citet{caron2020swav} use a global-local multi-crop scheme for self-supervised learning, and \citet{caron2021dino} show that multi-crop training with ViTs yields attention maps that segment objects. These methods still crop from the image alone, without using the text query. For retrieval, LARE~\citep{alquwayfili2026lare} places crops on the low-attention regions of a frozen encoder and combines their embeddings with the global image embedding at inference time. MINER's crops are query-agnostic and geometrically fixed (the center and four corners); we evaluate content-aware placement only as an ablation (Section~\ref{sec:results:analysis}) and find it adds nothing over fixed coverage.

A separate line studies which attention coefficients carry the
matching signal. Attention rollout~\citep{abnar2020quantifying} and
relevance propagation~\citep{chefer2021generic} approximate the
information flow to the output token, and
Grad-ECLIP~\citep{zhao2025gradeclip} identifies the CLS-row attention as
the slice that enters the score. Since attention heads are largely
redundant~\citep{voita2019analyzing}, we average this CLS- or probe-row
slice over heads and use it as the encoder's own attention map in our
saliency-source ablation (Section~\ref{sec:results:analysis}), where it
is one of several interchangeable options.

\subsection{Hubness Correction in Cross-Modal Retrieval}

In high-dimensional embedding spaces, certain points become universal
nearest neighbours (``hubs''), distorting nearest-neighbour
retrieval~\citep{radovanovic2010hubs}. Cross-domain Similarity Local
Scaling (CSLS), originally proposed for unsupervised word
translation~\citep{conneau2018word}, corrects for this by subtracting a
query-side and a gallery-side $k$-nearest-neighbour mean similarity
from each score. In cross-modal retrieval,
\citet{bogolin2022crossmodal} propose Querybank
Normalisation (QBNorm) and \citet{wang2023dbnorm} propose
DBNorm, both rescaling similarities against a bank of samples;
\citet{onesinglehub2026} further show that a single hub
text can score highly against many unrelated images in modern
CLIP-based retrieval. Hubness corrections beyond bank-based rescaling
include the Inverted Softmax~\citep{smith2017offline}, which turns
similarities into softmax-normalised retrieval probabilities but
requires an inverse temperature tuned on held-out data, and Mutual
Proximity~\citep{schnitzer2012local}, which rescales each distance into
the probability that two points are mutually close, in practice by
fitting a parametric model to the distance distribution. Two-sided CSLS
needs neither: subtracting local neighbourhood means makes the
correction invariant to the scale of the similarity scores and
introduces only a neighbourhood size $k$ to which performance is
insensitive. We therefore apply two-sided CSLS at inference time
(Section~\ref{sec:method:csls}) and compare it empirically against
QBNorm, a natural drop-in alternative, in
Section~\ref{sec:results:analysis}.

%% file: sec/3_dataset.tex
\section{Rare Objects in Cluttered Scenes Dataset}
\label{sec:rocs}
\begin{figure}[htbp]
\centering
\includegraphics[width=1.05\textwidth]{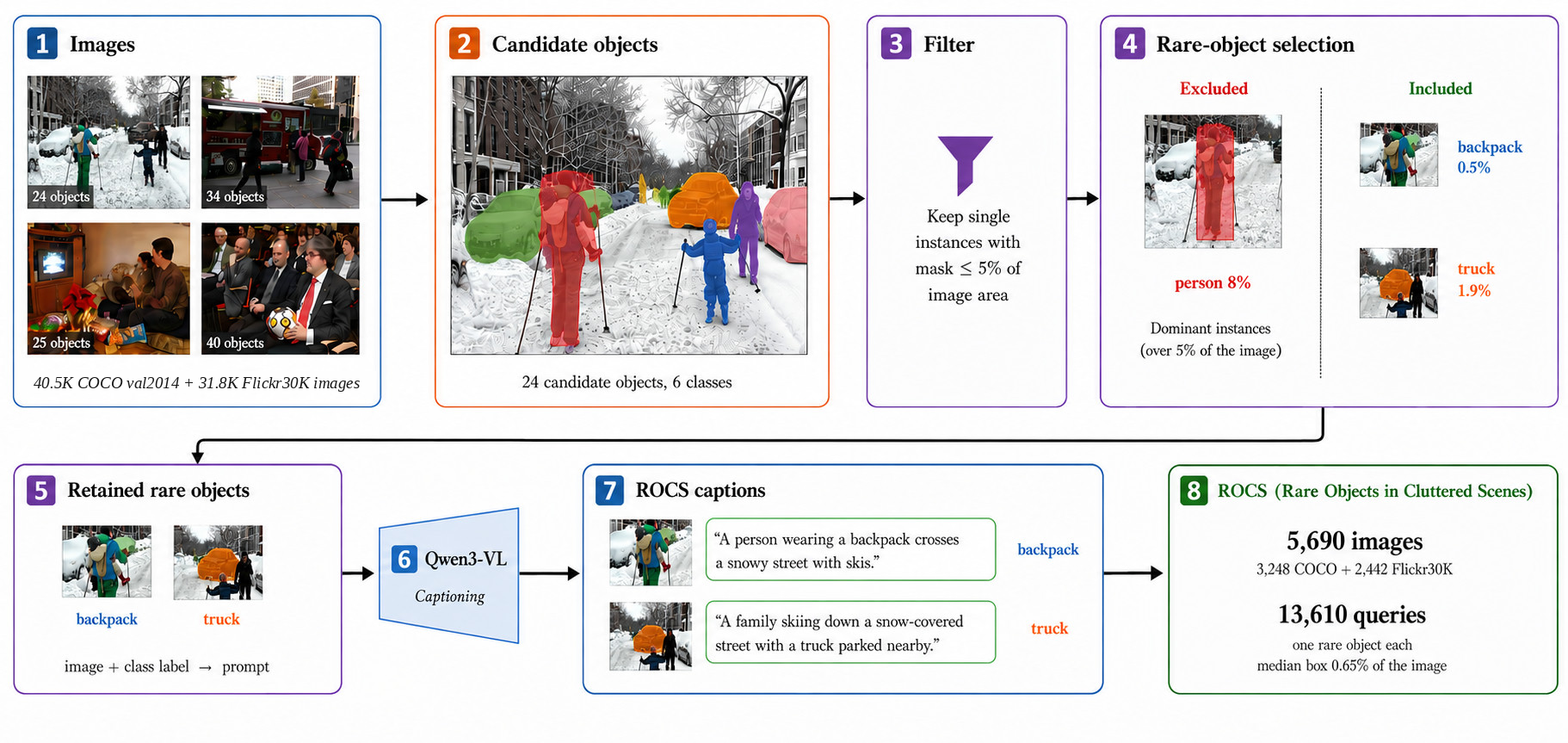}
\figtightcap
\caption{ROCS curation pipeline: SAM3 detection with clutter ranking (top $10\%$ kept), rare-object selection (single-instance classes with masks of at most $5\%$ of the image; dominant instances pruned), and Qwen3-VL captioning anchored on each retained object.}
\label{fig:rocs-pipeline}
\end{figure}

To evaluate image retrieval in crowded scenes, we construct \textbf{ROCS}
(Rare Objects in Cluttered Scenes), a benchmark of images with many
object instances and underrepresented classes, built by the automated
pipeline in Figure~\ref{fig:rocs-pipeline}.

\begin{figure}[t]
\centering
\includegraphics[width=0.9\linewidth]{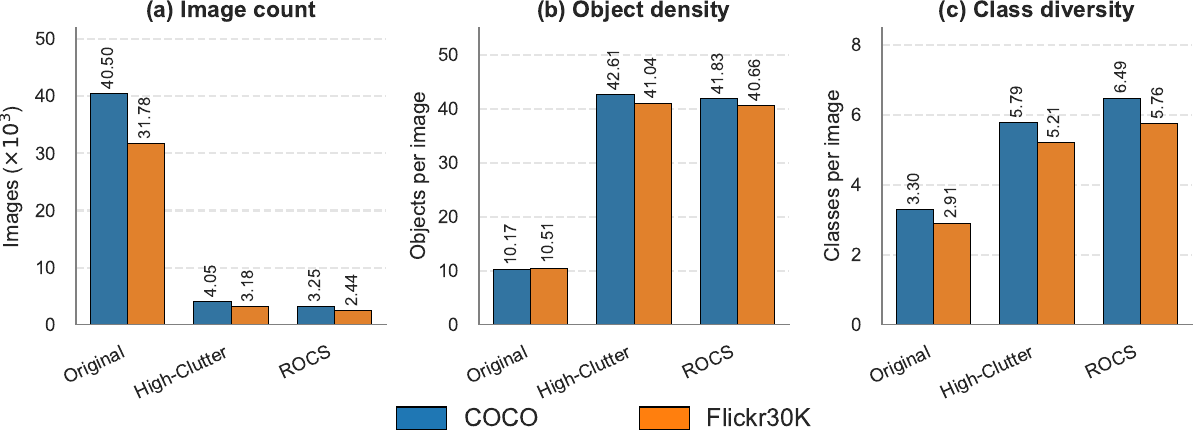}
\figtightcap
\caption{ROCS stage-wise statistics.
(a) Number of images retained at each stage (thousands).
(b) Object density jumps by ${\sim}4{\times}$ in the High-Clutter step
on both splits.
(c) The rare-class filter that produces ROCS additionally broadens
per-image class diversity.}
\label{fig:rocs-stats}
\end{figure}

\subsection{Object Detection}
\label{sec:rocs-construction}
The first stage of the pipeline focuses on identifying heavily
cluttered images. We begin by processing all images from the
COCO~\citep{lin2014microsoft} and
Flickr30K~\citep{plummer2015flickr30k} (all images of COCO val2014 and of Flickr30K) using
SAM3~\citep{carion2026sam3}, a promptable segmentation model. To ensure
consistency with established vocabulary, we prompt SAM3 with the
canonical list of object categories used in MS COCO captions; the six
rephrased prompts are listed in the supplementary material. For each
image, SAM3 outputs a set of detected object instances together with
their predicted class labels and bounding boxes, from which we compute
three image-level statistics: $(i)$ the total number of segmented
object instances, $(ii)$ the number of unique object categories, and
$(iii)$ per-class instance frequencies.

To construct the cluttered candidate pool, images are ranked in
descending order by total object count, and the top \textbf{10\%} are
selected. This yields the \emph{High-Clutter Subset} that serves as
input to the next stage.

\subsection{Rare Object Selection}
\label{sec:rocs-rare-selection}
Within the High-Clutter Subset, we identify \emph{rare classes} at the
image level, defined as object categories that appear exactly once in
a given image. In crowded scenes, such single-instance categories
often correspond to small or low-salience objects that are easily
overlooked by global representations. For each rare-class instance, we
prompt SAM3 a second time with its bounding box to extract an
instance-level segmentation mask, yielding per-instance mask
annotations for every rare object.

\begin{figure}[H]
\centering
\includegraphics[width=0.85\linewidth]{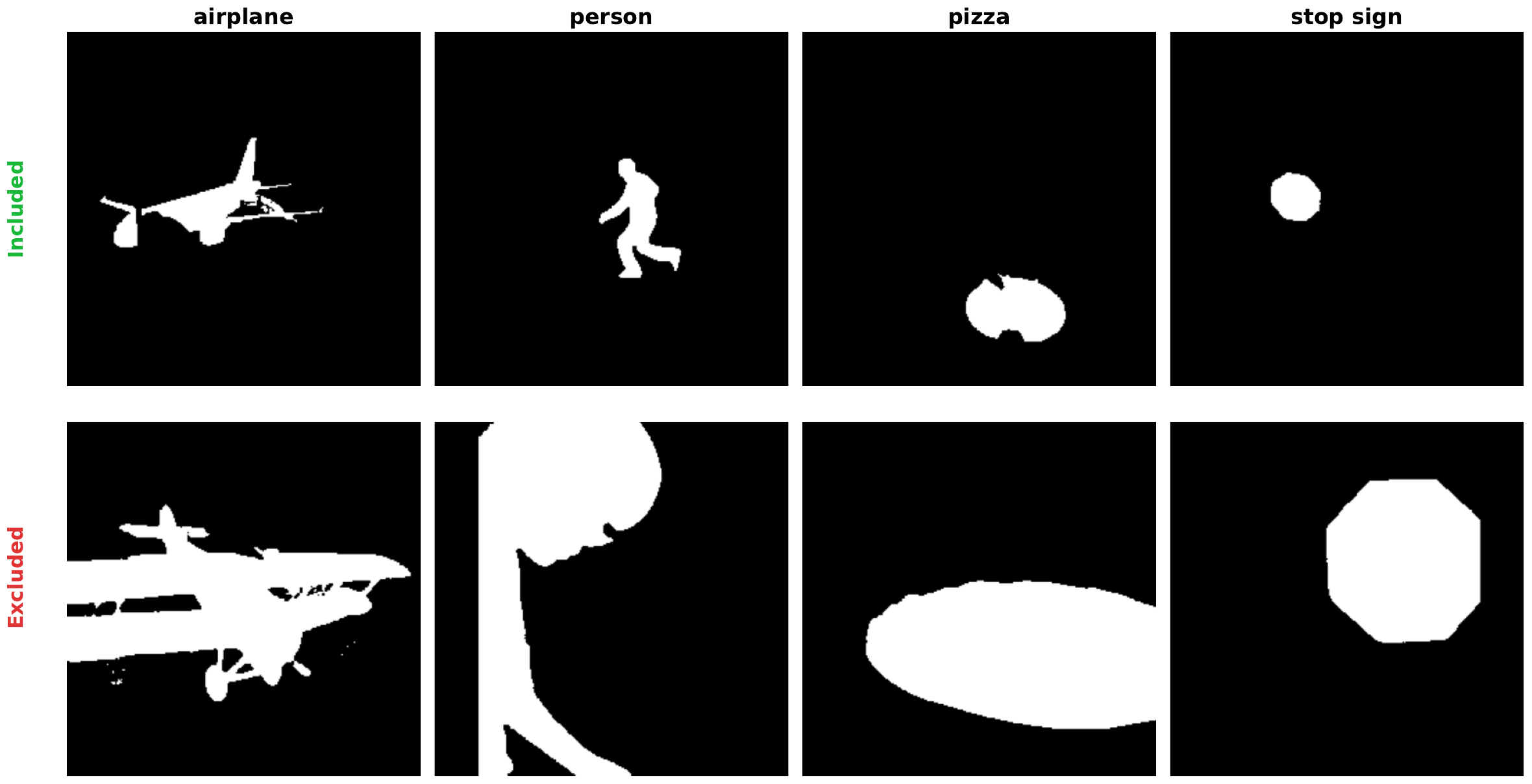}
\figtightcap
\caption{Rare object selection. For each rare-class instance we
measure the SAM3 mask area as a fraction of the image area. Instances
covering at most $5\%$ of the image (top row, \emph{included}) are
kept as low-salience targets; instances of the same class but covering
more than $5\%$ (bottom row, \emph{excluded}) are dropped as visually
dominant.}
\label{fig:prominence_filter}
\end{figure}

For each rare-class mask, we discard any instance whose mask area
exceeds \textbf{5\%} of the total image area
(Figure~\ref{fig:prominence_filter}); such instances are visually
dominant rather than low-salience. SAM3's masks give a tighter estimate
of visual footprint than bounding boxes, which overstate the area of
elongated or irregularly shaped objects. The final
ROCS subset consists of images from the High-Clutter Subset that
retain at least one rare-class instance after this filtering step.

Figure~\ref{fig:rocs-stats} reports image counts, average objects per
image, and average classes per image for the stages of
Figure~\ref{fig:rocs-pipeline}: the Original Test Set, the High-Clutter
Subset (top $10\%$ by object count), and the final ROCS after
rare-object selection. The final ROCS contains
substantially more objects and a broader set of categories than the
original splits.

\subsection{VLM Captioning}
\label{sec:rocs-vlm-captioning}
The final stage of the pipeline regenerates captions for the curated
ROCS images. The goal of this re-captioning step is to produce more
challenging textual descriptions that explicitly emphasize
low-attention regions, i.e., rare-class instances. In contrast, the
original dataset captions typically describe the dominant scene
context and often overlook small or underrepresented objects.

The rare-class labels surfaced by rare object selection are used as
guidance for a vision--language model. Specifically, we adopt
Qwen3-VL~\citep{qwen2025qwen3vl} and prompt it with class-aware
templates \textit{(e.g., ``Describe this scene. Focus on the [class]
that is visible.'')} to encourage explicit mention of these
underrepresented objects in the generated description; the exact
system and user prompts are given in the supplementary material. The model outputs one COCO-style caption per image.
\begin{samepage}
Each caption is a single sentence of roughly $12$ tokens that describes the
scene and names the rare object. Those objects are small: after the $5\%$
prominence filter of Section~\ref{sec:rocs-rare-selection}, the median rare
object's bounding box covers $0.65\%$ of the image and $57\%$ cover less
than $1\%$. A query
therefore turns on evidence occupying a fraction of the frame, which is
exactly what a single pooled image embedding represents least well.
A human audit of a random sample of queries is reported in the
supplementary material.
\end{samepage}

%% file: sec/4_method.tex
\section{Methodology}
\label{sec:method}

\begin{figure}[H]
\centering
\makebox[\textwidth][c]{\includegraphics[width=1.12\textwidth]{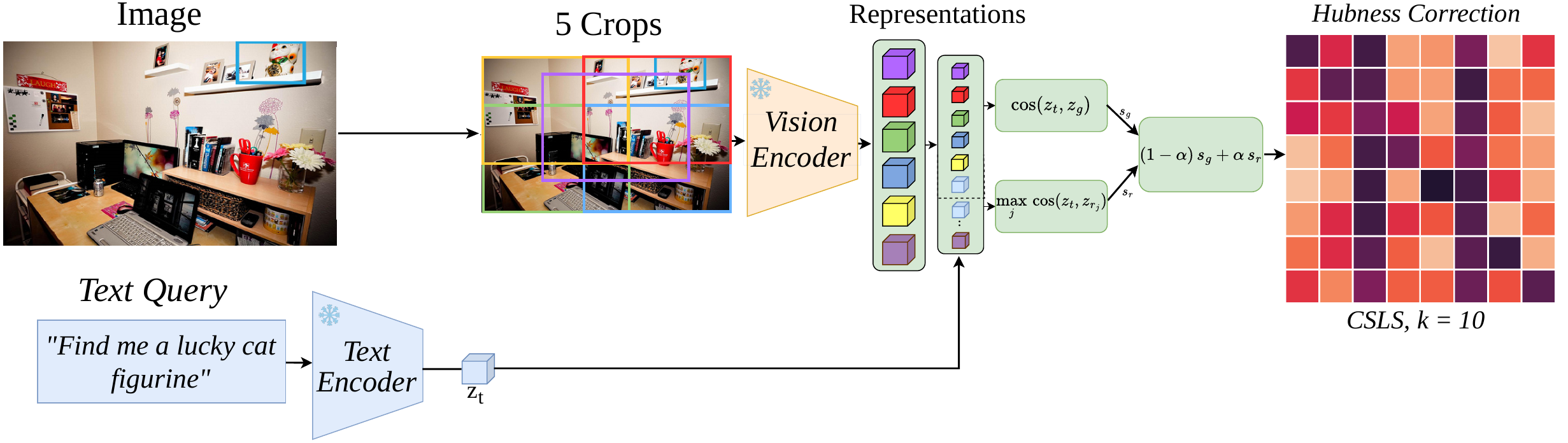}}
\figtightcap
\caption{\textbf{Overview of MINER.} Each image is encoded once by a frozen
dual encoder to obtain a global representation and a small set of
region-level embeddings extracted from a predefined five-crop layout. The
global and regional similarities are fused and subsequently refined using
two-sided CSLS, yielding improved text-to-image retrieval without retraining
or query-conditioned inference.}
\label{fig:pipeline}
\end{figure}

MINER is fully training-free, built on a frozen dual-encoder
backbone (Figure~\ref{fig:pipeline}). It has two inference-time stages.
First, we augment each image's global embedding with a small bank of
\emph{fixed} regional crops and fuse the global and regional similarities
into a single score. Second, we rescore the similarity matrix with
two-sided CSLS to correct hubness in the joint embedding space. Both
global and regional features come from the same frozen encoder; nothing
is trained or tuned per encoder.

\subsection{Region Candidates}
\label{sec:method:attention}

A frozen dual encoder exposes a single image vector that is directly
comparable to text: the projected CLS token in CLIP and the
\textsc{AttentionPoolLatent} output in SigLIP-family
models~\citep{tschannen2025siglip}. Patch-level tokens bypass the
contrastive projection head and cannot be matched to text
directly, so we obtain regional evidence by re-encoding image crops
through the same head rather than by reading patch tokens.

We use $N{=}5$ fixed crops at $60\%$ scale: the center and the four
corners, each $0.6H \times 0.6W$, so a crop keeps the image's aspect
ratio and covers $36\%$ of its area. These crops are parameter-free and
retrieve better than the alternatives (Table~\ref{tab:cropgen}). This choice is
deliberate: crop placement barely matters provided coverage is spread
out, so the fixed set matches a $3{\times}3$ grid and the saliency-guided
variants of Section~\ref{sec:results:analysis}, while purely
attention-placed and random crops lag (Table~\ref{tab:cropgen}). Each region $r$ is
encoded by the same frozen backbone as the global image,
$\mathbf{z}_r = f_v(r)$, with crops bicubically resized to the encoder's
native input size.

\begin{table}[htbp]
\centering
\footnotesize
\setlength{\tabcolsep}{8pt}
\caption{Crop-placement strategies, R@1 on SigLIP\,2\,So/16, all rows hubness-corrected.}
\label{tab:cropgen}
\begin{tabular}{@{}lcc@{}}
\toprule
Crop-placement strategy & ROCS-COCO & ROCS-Flickr30K \\
\midrule
Baseline (no crops) & 50.29 & 52.28 \\
\midrule
Fixed 5-crop        & 52.36 & 53.84 \\
$3{\times}3$ grid    & 51.96 & 53.69 \\
Random              & 51.06 & 52.84 \\
5 attention crops   & 51.40 & 53.13 \\
\bottomrule
\end{tabular}
\end{table}

\subsection{Region Fusion}
\label{sec:method:augment}

Let $s_g = \cos(\mathbf{z}_t, \mathbf{z}_g)$ be the global similarity
and $s_r = \max_{j} \cos(\mathbf{z}_t, \mathbf{z}_{r_j})$ the
strongest regional similarity, where $\mathbf{z}_g = f_v(I)$ is the
global image embedding and $\mathbf{z}_t$ the text embedding from the
frozen text encoder. We blend the two:
\begin{equation}
S(\alpha) = (1 - \alpha)\, s_g + \alpha\, s_r,
\label{eq:blend}
\end{equation}
where $\alpha \in [0, 1]$ is the blend weight, $j \in \{1, \dots, N\}$
indexes the crop embeddings $\mathbf{z}_{r_j}$ of
Section~\ref{sec:method:attention}, and $S(\alpha)$ is the fused
text--image score. We use $\alpha = 0.4$ as the default. The optimum is broad:
$\alpha \in [0.2, 0.6]$ is within $1$ R@1 of the peak on
ROCS-COCO (see the supplementary material).

Equation~\ref{eq:blend} also explains where the gain comes from: the
fused score is the global similarity plus $\alpha$ times the excess of
the best crop over it. For the correct image the crop containing the
named object recovers what global pooling dilutes for a small region;
for a wrong image no crop typically contains the object, so the excess is
noise. MINER thus flips a query when $\alpha$ times that excess outweighs
the baseline margin, which predicts gains concentrating on small objects,
placement barely mattering, and the gain vanishing when the object is
masked (Section~\ref{sec:results:analysis}). A confidence-gated variant that blends only when $s_g < \tau$ and $s_r > s_g$ fires on ${\sim}5\%$ of pairs on COCO\,5K and yields $\leq +0.04$ R@1, so Equation~\ref{eq:blend} applies to every pair.

\subsection{Hubness Correction}
\label{sec:method:csls}

Let $\mathbf{S}$ be the fused similarity matrix, whose entry
$\mathbf{S}_{t, i}$ is the fused score $S(\alpha)$ between text query
$t$ and gallery image $i$. $\mathbf{S}$ is rescored with two-sided
CSLS~\citep{conneau2018word} to correct for hubness in the joint
embedding space~\citep{radovanovic2010hubs}:
\begin{equation}
\mathrm{CSLS}(t, i) = 2\,\mathbf{S}_{t, i}
- \frac{1}{k}\!\sum_{j \in \mathcal{N}_k^{I}(t)} \mathbf{S}_{t, j}
- \frac{1}{k}\!\sum_{u \in \mathcal{N}_k^{T}(i)} \mathbf{S}_{u, i},
\label{eq:csls}
\end{equation}
where $\mathcal{N}_k^{I}(t)$ is the set of the $k$ gallery images with
the highest similarity to query $t$ under $\mathbf{S}$,
$\mathcal{N}_k^{T}(i)$ is the set of the $k$ queries with the highest
similarity to image $i$, and $k = 10$. The two correction terms subtract each query's mean
similarity to its $k$ nearest gallery items and each item's mean
similarity to its $k$ nearest queries, penalizing universal nearest
neighbors. Both terms are computed once on the full test-split
similarity matrix (all captions of the split against all its images),
with no batching and no external query bank; the correction is thus
transductive within the evaluated split.

The supplementary material shows that performance is flat across
$k \in \{5, 10, 20\}$; $k{=}1$ underperforms by ${\sim}1$ R@1. CSLS
and region augmentation are near-additive: on ROCS-COCO,
the two single-axis lifts compose to within ${\sim}1$ R@1 of their
joint lift (Section~\ref{sec:results:analysis}).

%% file: sec/5_experiments.tex
\section{Experiments}
\label{sec:results}

\subsection{Experimental Setup}
\label{sec:results:setup}

\paragraph{Backbones and metrics:}
We evaluate MINER on three frozen vision--language dual encoders representing two model families: CLIP\,L/14~\citep{radford2021learning}, SigLIP\,So/14~\citep{zhai2023siglip}, and SigLIP\,2\,So/16~\citep{tschannen2025siglip}. Following standard retrieval protocols, we report zero-shot text-to-image Recall@$K$ (\textit{R@1}, \textit{R@5}, and \textit{R@10}).

\paragraph{Datasets:}
We evaluate on four retrieval benchmarks. Standard COCO~5K (Karpathy split) and Flickr30K measure general retrieval; for cluttered scenes we add {ROCS-COCO} ($3{,}248$ images) and {ROCS-Flickr30K} ($2{,}442$ images), introduced in Section~\ref{sec:rocs}. These benchmarks are derived from high-clutter subsets of COCO and Flickr30K and re-captioned with Qwen3-VL~\citep{qwen2025qwen3vl} to emphasize a single rare or visually subordinate object in each image.

\paragraph{Implementation details:}
Unless otherwise specified, our method uses SigLIP\,2\,So/16 as the frozen backbone with five fixed region crops (center and four corners) at a crop ratio of $r=0.6$. Global and region similarities are combined using a blending weight of $\alpha=0.4$, followed by two-sided CSLS rescoring with $k=10$. MINER is training-free at inference.  Table~\ref{tab:cropgen} compares crop-placement strategies; the supplementary material reports the hyperparameter sweeps and the saliency-source comparison.

\subsection{Main Results}
\label{sec:results:main}

\begin{figure}[t]
\centering
% WACV figure used at its native size; jmlr's column is wider, so a
% fraction rather than \linewidth keeps the cells at the intended scale.
\includegraphics[width=\linewidth]{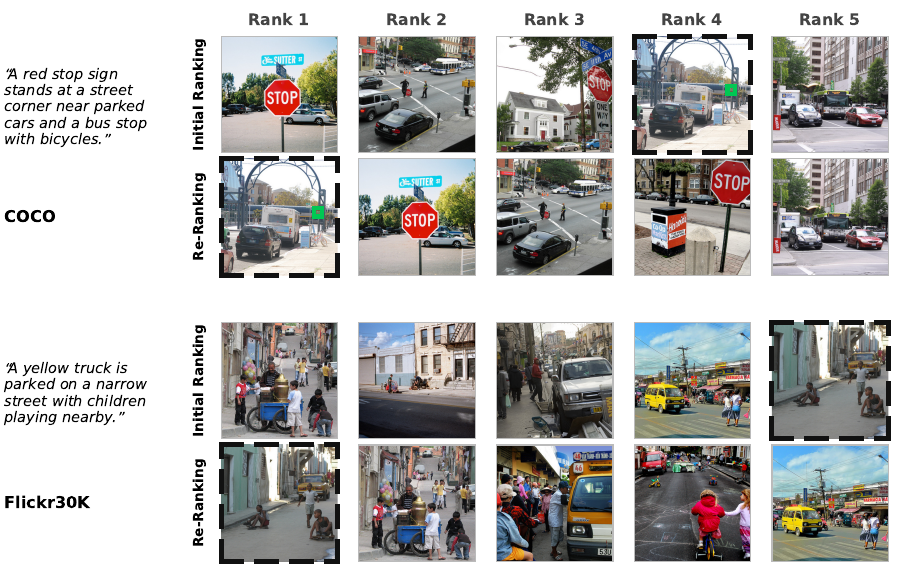}
\caption{\textbf{Qualitative re-ranking on ROCS.} For one COCO query
(top) and one Flickr30K query (bottom), each naming a rare object, we show
the top-5 gallery images under the baseline global retrieval
(\emph{Initial Ranking}) and after MINER's region augmentation and
rescoring (\emph{Re-Ranking}). The ground-truth image is outlined with a
dashed box wherever it appears, and the named rare object is boxed in green
on COCO. In both cases the ground truth sits low under the baseline, which
is dominated by the surrounding scene, and MINER lifts it to rank~1.}
\label{fig:visual-results}
\end{figure}

Table~\ref{tab:main} shows that MINER improves R@1 on every backbone and
split, with no regressions. The gain is largest where the query hinges
on a small object ($+5.28$ on ROCS-COCO and $+5.84$ on ROCS-Flickr30K
with SigLIP\,2, against $+3.69$ and $+3.36$ on the standard splits) and
on the weaker backbones (up to $+9.66$ for CLIP\,L/14).
Figure~\ref{fig:visual-results} shows a representative example; the
ablation below shows that the two stages contribute complementary,
near-additive gains. Image-to-text retrieval gains on the ROCS splits and is
neutral on the standard ones (supplementary material).

% Main results table (dense-first ordering: ROCS splits, then standard).
\begin{table*}[t]
\centering
\caption{Text-to-image retrieval across three frozen backbones and
four splits.}
\label{tab:main}
\setlength{\tabcolsep}{3.5pt}
\resizebox{\textwidth}{!}{%
\footnotesize
\begin{tabular}{ll ccc ccc ccc ccc}
\toprule
\textbf{Method} & \textbf{Backbone}
 & \multicolumn{3}{c}{\textbf{ROCS-COCO}}
 & \multicolumn{3}{c}{\textbf{ROCS-Flickr30K}}
 & \multicolumn{3}{c}{\textbf{COCO 5K}}
 & \multicolumn{3}{c}{\textbf{Flickr30K}} \\
\cmidrule(lr){3-5}\cmidrule(lr){6-8}\cmidrule(lr){9-11}\cmidrule(lr){12-14}
 & & R@1 & R@5 & R@10 & R@1 & R@5 & R@10 & R@1 & R@5 & R@10 & R@1 & R@5 & R@10 \\
\midrule
Baseline                       & CLIP\,L/14        & 29.10 & 49.65 & 58.38 & 31.98 & 51.72 & 61.07 & 36.32 & 61.08 & 71.11 & 64.48 & 87.12 & 91.92 \\
\textbf{MINER}                 & CLIP\,L/14        & 37.12 & 58.62 & 67.20 & 39.47 & 61.48 & 69.77 & 44.44 & 68.95 & 78.34 & 74.14 & 93.30 & 96.84 \\
\midrule
Baseline                       & SigLIP\,So/14     & 45.41 & 64.80 & 72.23 & 46.18 & 66.57 & 74.23 & 54.24 & 76.78 & 84.21 & 82.94 & 96.08 & 98.00 \\
\textbf{MINER}                 & SigLIP\,So/14     & 50.59 & 69.74 & 76.61 & 52.15 & 72.93 & 79.51 & 58.22 & 80.15 & 86.87 & 86.90 & 97.30 & 98.72 \\
\midrule
Baseline                       & SigLIP\,2\,So/16  & 47.08 & 66.60 & 74.18 & 48.00 & 68.45 & 75.98 & 56.55 & 78.75 & 85.95 & 83.72 & 96.34 & 98.32 \\
\textbf{MINER}                 & SigLIP\,2\,So/16  & \textbf{52.36} & \textbf{71.07} & \textbf{78.01} & \textbf{53.84} & \textbf{73.42} & \textbf{80.35} & \textbf{60.24} & \textbf{81.32} & \textbf{87.98} & \textbf{87.08} & \textbf{97.58} & \textbf{98.78} \\
\bottomrule
\end{tabular}}
\end{table*}

\subsection{Ablation Study}
\label{sec:results:analysis}

We ablate the design choices behind MINER; full tables are in the supplementary material.

\paragraph{Stage contributions.}
The two stages are near-additive: on ROCS-COCO the rescoring alone adds
$+3.21$ R@1 and the crops alone $+2.03$, versus $+5.28$ together. The
pattern holds on all three backbones and both ROCS splits.

\paragraph{Caption specificity.}
ROCS images come from the same pools as the standard splits, but each
ROCS caption names one low-salience object. With rescoring fixed, the
region stage adds $+0.7$ to $+0.8$ R@1 on the standard splits and $+1.6$
to $+2.1$ on the ROCS splits: region augmentation pays off most when the
query hinges on an object the global embedding underweights.

\begin{figure}[t]
\centering
\includegraphics[width=\linewidth]{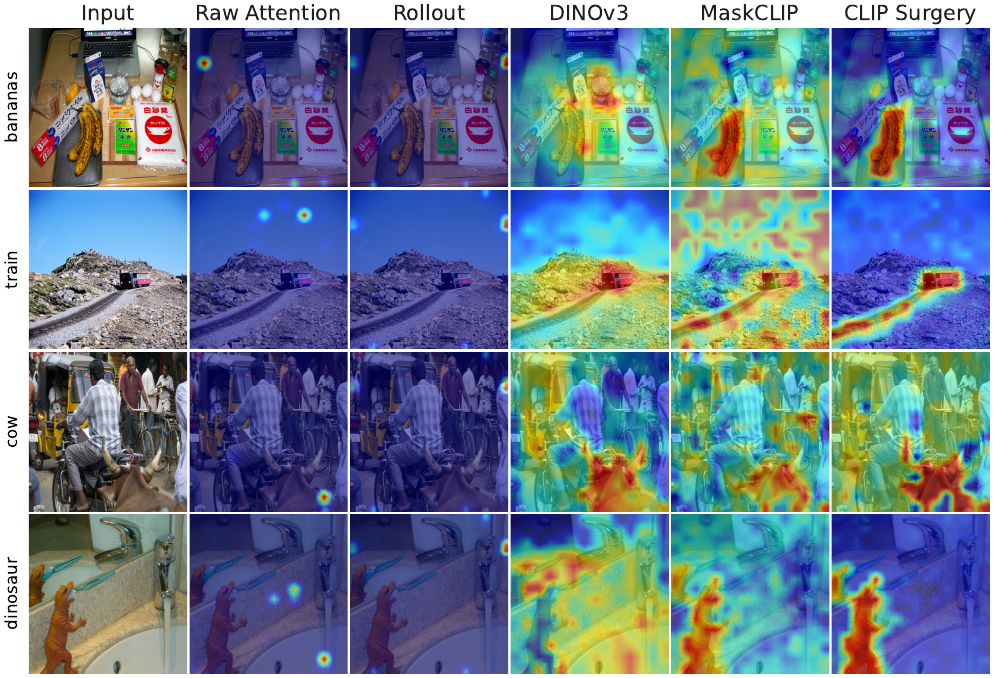}
\figtightcap
\caption{Saliency maps for four COCO val2014 images (rows) under five
saliency sources (columns, after the input; attention rollout is shown
for visual comparison only). The sources produce visibly different maps,
yet all place the guided crop such that retrieval stays within $1$ R@1 of
the parameter-free fixed crops (supplementary material): what the crop
covers matters, not which saliency map selects it.}
\label{fig:saliency-comparison}
\end{figure}

\paragraph{Coverage.}
The region gain is governed by how much of the image the crops cover, not by where they sit. The crop-placement strategy barely
matters (Table~\ref{tab:cropgen}): a parameter-free fixed 5-crop set
matches a $3{\times}3$ grid and outperforms placing all five crops on
attention peaks, while random, low-coverage crops lag furthest. The saliency source is equally immaterial: maps as different as the encoder's own attention, MaskCLIP~\citep{zhou2022maskclip}, DINOv3~\citep{simeoni2025dinov3}, and CLIP-Surgery~\citep{li2023clipsurgery} all keep R@1 within a point of the fixed crops (Figure~\ref{fig:saliency-comparison}). Rerankers that need a cross-encoder or a detector fall outside the training-free, single-encoder setting studied here.

\paragraph{Pooling mechanism.}
The pattern points to the SigLIP-family \textsc{AttentionPoolLatent}
layer responding to how much of the patch grid a crop covers, not to
where it sits. A direct patch-masking probe confirms this: at equal patch count a \emph{random} subset largely preserves the matched image--text cosine, whereas a \emph{contiguous} quadrant degrades it (details in the supplementary material). This is the
failure mode MINER's spread-out crops avoid.

\paragraph{Target masking.}
To check that the gain comes from the named object, we mask its SAM3 box
in the ground-truth image, re-encode, and re-rank. On the ROCS-COCO
queries that MINER fixes over the corrected baseline, the correct image
loses rank~1 in $53\%$ of cases, against $16\%$ when a random region of
the same size is masked; the gap holds for objects under $0.5\%$ of the
image and on the other two backbones, and the highest-scoring crop
contains the object $68\%$ of the time against $29\%$ chance.

\paragraph{Dense patch matching.}
The same pooling property explains why we crop rather than match dense
patch descriptors, the training-free alternative of MaskCLIP-style
methods~\citep{zhou2022maskclip}. Projecting each SigLIP\,2 patch token
through the pooling head and retrieving by late interaction reaches only
$29.1$ R@1 on ROCS-COCO, far below the global baseline ($47.1$) and
MINER ($52.4$): the cross-modal alignment of these encoders lives in the
pooled representation, which a crop preserves and a patch does not.

\paragraph{Hubness.}
The rescoring stage works because these frozen encoders are measurably
hub-afflicted: on ROCS-COCO with SigLIP\,2, two-sided CSLS lowers the
$k$-occurrence skewness from $2.35$ to $1.80$ and cuts gallery images that
are never retrieved from $1.6\%$ to $0.4\%$ (supplementary material).

\paragraph{Alternative corrections.}
Our protocol computes both CSLS terms on the evaluated split. Estimating
the gallery-side term from a disjoint query bank instead costs almost
nothing ($+2.65$ versus $+2.80$ R@1), so MINER does not depend on seeing
the test queries together. Against the strongest alternatives on the same
protocol (Table~\ref{tab:hubness-baselines}), CSLS is the best untuned
correction: NNN~\citep{chen2024nnn} and DBNorm~\citep{wang2023dbnorm}
trail it on ROCS-COCO and edge it by $0.26$ on ROCS-Flickr30K; Sinkhorn
normalization wins on one split at a hand-picked temperature but swings
by up to $8$ R@1 across temperatures; QBNorm~\citep{bogolin2022crossmodal}
underperforms at every temperature and collapses at its default
$\beta{=}20$, since its $\exp(\beta\cos)$ weighting assumes a similarity
range that frozen encoders do not produce.

\begin{table}[htbp]
\centering
\footnotesize
\setlength{\tabcolsep}{5pt}
\caption{Hubness corrections on held-out queries, R@1 with region fusion; alternatives at their best swept setting, CSLS at $k{=}10$.}
\label{tab:hubness-baselines}
\begin{tabular}{@{}lcc@{}}
\toprule
Correction & ROCS-COCO & ROCS-Flickr30K \\
\midrule
None                          & 48.71 & 49.00 \\
DBNorm (best)                 & 51.00 & 52.08 \\
NNN (best)                    & 51.38 & 52.08 \\
Sinkhorn ($\tau{=}0.02$, converged) & 50.66 & \textbf{52.68} \\
CSLS ($k{=}10$)               & \textbf{51.51} & 51.82 \\
\bottomrule
\end{tabular}
\end{table}

\paragraph{Hyperparameters.}
All four hyperparameters have broad optima (supplementary material): $\alpha$ is within $1$ R@1 of its peak across
$[0.2, 0.6]$, $r$ is flat across $[0.5, 0.7]$, recall rises with $N$ to
the structural cap of $5$, and $k$ is flat across $\{5, 10, 20\}$. We use
$\alpha{=}0.4$, $r{=}0.6$, $N{=}5$, $k{=}10$ throughout.

\subsection{Inference Cost and Scalability}
\label{sec:efficiency}

\begin{table}[H]
\centering
\footnotesize
\setlength{\tabcolsep}{4pt}
\caption{Per-image encoding latency on SigLIP\,2\,So/16 over ROCS-COCO.}
\label{tab:latency}
\begin{tabular}{lrr}
\toprule
Method & Passes & ms / img \\
\midrule
Baseline            & 1  & 99  \\
Random crops        & 6  & 471 \\
Attention-based     & 6  & 466 \\
Fixed 5-crop (MINER) & 6 & 471 \\
Uniform $3 \times 3$ & 10 & 773 \\
\bottomrule
\end{tabular}
\end{table}

Region augmentation adds one encoder pass per crop (Table~\ref{tab:latency}; single Quadro RTX\,8000, batch size $1$):
at $N{=}5$ MINER costs ${\sim}4.8{\times}$ the global pass, against
$7.8{\times}$ for a $3{\times}3$ grid. The crops are encoded once at
indexing time and raise gallery storage $6\times$ (one global plus five
crop embeddings per image; at $D{=}1152$ in fp16, $2.3$\,GB per million
images for the global index against $14$\,GB with crops). Online,
blending and rescoring add $0.32$\,ms per query. Rescoring only a
top-$M$ shortlist removes the crop index while keeping the full gain
(within $0.15$ R@1 at $M{=}25$, $76$\,ms per query). When the full query
set is not available in advance, the gallery-side CSLS term is
precomputed from a representative query bank and the query-side term
over the top-$M$ candidates; with $M{=}25$ this is never more than $0.4$
R@1 below the transductive setting on any backbone and split, and up to
$0.8$ above it, so it is the configuration we recommend for deployment.
Growing the gallery to $11{,}232$ images with distractors, MINER stays
ahead of the baseline on all $24$ backbone, query-set and size
combinations, with the region gain at $+1.5$ to $+2.4$ R@1 throughout.
All four studies are in the supplementary material.
\FloatBarrier

%% file: sec/6_conclusion.tex
\section{Conclusion}
\label{sec:conclusion}

We presented MINER, a training-free inference pipeline that augments a
frozen dual encoder's global embedding with a bank of fixed regional
crops and rescores the resulting similarities to correct hubness. We
also presented ROCS, a benchmark of cluttered scenes whose captions
each name a rare object. MINER improves R@1 on every backbone and split we tested, with
the largest gains where global pooling fails most: rare-object queries
and weaker encoders. Two analyses explain the design. First, the two
stages are complementary and near-additive: the crops recover localized
evidence the global embedding underweights, while the rescoring removes
measurable hubness in the joint space. Second, what region augmentation adds is
governed by spatial coverage, not localization: a parameter-free set of
fixed crops matches every saliency-guided variant, and the effect traces
to the encoder's pooling responding to how much of the patch grid a crop
covers. MINER is therefore a simple drop-in for existing dual-encoder
retrieval systems.

%% file: sec/appendix_algorithm.tex
\section{Inference Algorithm}
\label{app:algorithm}

Algorithm~\ref{alg:pipeline} summarises the full training-free inference
pipeline described in the main paper. Gallery embeddings (the global
embedding and the $N$ regional embeddings of every image) are computed
once at indexing time and reused across all queries; only the fusion and
CSLS rescoring depend on the query. The same frozen encoder $f_v$ is used
for the global image and for every crop, and the text encoder $f_t$ is
never modified.

\begin{algorithm2e}[H]
\SetAlgoLined
\DontPrintSemicolon
\KwIn{Query texts $\{t_u\}_{u=1}^{Q}$; gallery images $\{i_m\}_{m=1}^{M}$;
frozen encoders $f_v, f_t$; blend $\alpha$, crops $N$, ratio $r$,
neighbourhood $k$.}
\KwOut{Ranking of gallery images for each query.}
\BlankLine
\tcp{Indexing: once per gallery image, query-independent}
\For{each image $i_m$}{
  $\mathbf{z}_g^{(m)} \leftarrow f_v(i_m)$ \tcp*{global embedding}
  $\mathcal{R} \leftarrow$ center crop $\cup$ 4 corner crops, each $r H \times r W$ (aspect-preserving, $r^2$ of the area)\;
  \For{each region $\rho \in \mathcal{R}$}{
    $\mathbf{z}_{\rho}^{(m)} \leftarrow f_v(\rho)$ \tcp*{regional embedding}
  }
}
\BlankLine
\tcp{Retrieval: per query}
\For{each query $t_u$}{
  $\mathbf{z}_t \leftarrow f_t(t_u)$\;
  \For{each image $i_m$}{
    $s_g \leftarrow \cos(\mathbf{z}_t, \mathbf{z}_g^{(m)})$\;
    $s_r \leftarrow \max_{\rho}\cos(\mathbf{z}_t, \mathbf{z}_{\rho}^{(m)})$\;
    $\mathbf{S}_{u,m} \leftarrow (1-\alpha)\,s_g + \alpha\,s_r$ \tcp*{region-global blend}
  }
}
$\mathbf{S} \leftarrow \mathrm{CSLS}(\mathbf{S}; k)$ \tcp*{two-sided CSLS rescoring}
\Return{$\mathrm{argsort}_m\,\mathbf{S}_{u,m}$ for each query $t_u$}\;
\caption{Training-free region-augmented retrieval with hubness correction.}
\label{alg:pipeline}
\end{algorithm2e}

%% file: sec/appendix_sweeps.tex
\section{Full Hyperparameter Sweeps}
\label{app:sweeps}

\begin{figure}[h]
\centering
\includegraphics[width=\linewidth]{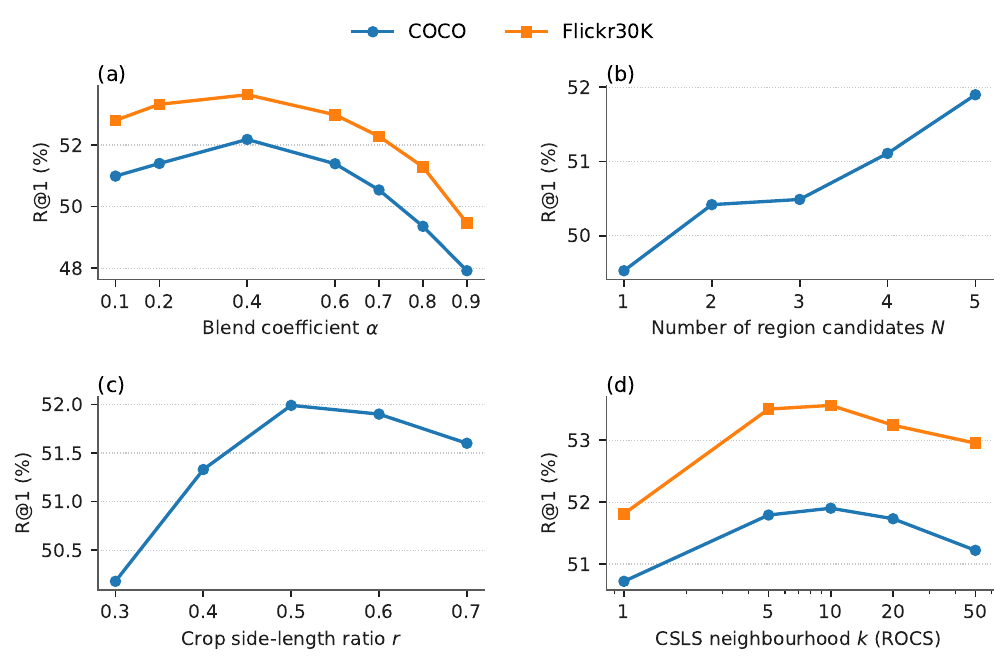}
\caption{Hyperparameter sweeps on the ROCS splits (SigLIP\,2\,So/16,
text-to-image R@1). Blend weight $\alpha$ and CSLS neighbourhood $k$ are
shown for both ROCS-COCO and ROCS-Flickr30K; the number of regions $N$
and crop ratio $r$ are shown on ROCS-COCO. Defaults: $\alpha{=}0.4$,
$N{=}5$, $r{=}0.6$, $k{=}10$.}
\label{fig:ablation-panels}
\end{figure}

Figure~\ref{fig:ablation-panels} plots the four hyperparameter sweeps
summarised in Section~5.3 of the main paper and extends the blend weight $\alpha$ and the CSLS neighbourhood $k$ to both
ROCS splits.
The trends are consistent across splits: the blend weight is within
$1$ R@1 of its peak across $\alpha \in [0.2, 0.6]$, recall rises
monotonically with the number of regions $N$ up to the structural cap of
$5$, the crop ratio $r$ is flat across $[0.5, 0.7]$, and the CSLS
neighbourhood $k$ is flat across $\{5, 10, 20\}$ with only $k{=}1$
underperforming. These broad optima are why a single default
($\alpha{=}0.4$, $N{=}5$, $r{=}0.6$, $k{=}10$) transfers across backbones
and splits without per-dataset tuning.

%% file: sec/appendix_pooling_probe.tex
\section{Patch-masking probe of the pooling layer}
\label{app:pooling}
To probe the SigLIP\,2 \textsc{AttentionPoolLatent} layer directly, we
mask subsets of the patch grid and measure the resulting pooled
image--text cosine similarity, averaged over the ROCS-COCO images.
Masking a \emph{random} $25\%$ of patches leaves the matched
similarity unchanged (mean cosine $0.139$, matching the full grid), and
even retaining only a \emph{random} $25\%$ preserves most of it
($0.126$), whereas restricting the input to a spatially
\emph{contiguous} $25\%$
subset drops it to $0.085$ (top-left quadrant); even a contiguous
\emph{half} of the grid (center $50\%$) scores worse ($0.100$) than a
scattered quarter. At equal patch
count, the pretrained pooling keeps its text alignment under
distributed coverage but not under a localized subset. This is a
property of the frozen encoder, not of our pipeline; it is the failure
mode that motivates spread-out crops. Table~\ref{tab:probe-full}
reports all masking conditions.

\begin{table}[h]
\centering
\footnotesize
\setlength{\tabcolsep}{10pt}
\caption{Patch-masking probe of the SigLIP\,2
\textsc{AttentionPoolLatent} layer, averaged over the ROCS-COCO images:
mean pooled image--text cosine of the matched caption, and cosine of
the pooled embedding to the full-grid pooled embedding. Scattered
subsets track the full grid; contiguous subsets do not.}
\label{tab:probe-full}
\begin{tabular}{@{}lcc@{}}
\toprule
Patch subset & Matched text cosine & Cosine to full grid \\
\midrule
Full grid                          & 0.139 & --    \\
Random $75\%$                      & 0.139 & 0.987 \\
Random $50\%$                      & 0.129 & 0.978 \\
Random $25\%$                      & 0.126 & 0.968 \\
Center $50\%$ (contiguous)         & 0.100 & 0.818 \\
Top-left quadrant ($25\%$, contiguous) & 0.085 & 0.879 \\
Single patch                       & $-0.029$ & 0.473 \\
\bottomrule
\end{tabular}
\end{table} This is consistent with the coverage
account in the main paper: a small attention-positioned crop is a
contiguous subset by construction and therefore degrades the pooled
embedding, which is why four spread-out corner crops are more reliable
than a single localized crop.

%% file: sec/appendix_detailed.tex
% Ported from the extended results developed after the ACML submission.
% Added for the camera-ready in response to the ACML 2026 reviews:
%   image-to-text results (R-Q8st), inductive CSLS (R-Q8st, R-Lba5),
%   top-M online re-ranking (R-Lba5), object-level attribution (R-Tz6A, R-Lba5).
\section{Additional Retrieval Results}
\label{app:additional}

\subsection{Image-to-Text Retrieval}
\label{app:i2t}

Table~\ref{tab:i2t} reports the reverse direction: each image ranks all
captions, and retrieval is correct if the top caption belongs to the
image. MINER lifts image-to-text R@1 on every ROCS cell ($+1.2$ to
$+2.3$) and is neutral on the standard splits ($-0.7$ to $+1.5$, with
Flickr30K already above $96$). The asymmetry is expected: crops add
image-side evidence for a rare object the caption names, whereas an
image query on the standard splits already matches on the dominant
scene.

\begin{table}[h]
\centering
\footnotesize
\setlength{\tabcolsep}{5pt}
\caption{Image-to-text R@1, corrected baseline vs MINER.}
\label{tab:i2t}
\begin{tabular}{@{}llccc@{}}
\toprule
Backbone & Split & Baseline & MINER & $\Delta$ \\
\midrule
\multirow{4}{*}{CLIP\,L/14}
 & ROCS-COCO   & 46.40 & \textbf{47.84} & $+1.44$ \\
 & ROCS-Flickr & 45.86 & \textbf{48.16} & $+2.30$ \\
 & COCO\,5K    & 62.10 & \textbf{63.18} & $+1.08$ \\
 & Flickr30K   & 88.10 & \textbf{89.60} & $+1.50$ \\
\midrule
\multirow{4}{*}{SigLIP\,So/14}
 & ROCS-COCO   & 59.91 & \textbf{61.21} & $+1.30$ \\
 & ROCS-Flickr & 58.89 & \textbf{61.14} & $+2.25$ \\
 & COCO\,5K    & \textbf{76.96} & 76.90 & $-0.06$ \\
 & Flickr30K   & 96.40 & \textbf{96.80} & $+0.40$ \\
\midrule
\multirow{4}{*}{SigLIP\,2\,So/16}
 & ROCS-COCO   & 60.99 & \textbf{62.50} & $+1.51$ \\
 & ROCS-Flickr & 59.99 & \textbf{61.18} & $+1.19$ \\
 & COCO\,5K    & 77.68 & \textbf{78.24} & $+0.56$ \\
 & Flickr30K   & \textbf{96.80} & 96.10 & $-0.70$ \\
\bottomrule
\end{tabular}
\end{table}

\subsection{Gain by Object Size}
\label{app:size}

Table~\ref{tab:size} bins ROCS-COCO queries by the relative image area of
the named rare object and reports R@1 over the hubness-corrected
baseline. The gain is present across the range and peaks for objects
covering $2$ to $5\%$ of the image; for the smallest objects it is not
statistically distinguishable from zero, which bounds what a
$60\%$-scale crop can recover.

\begin{table}[h]
\centering
\footnotesize
\setlength{\tabcolsep}{5pt}
\caption{R@1 gain over the corrected baseline by rare-object
bounding-box area, with paired-bootstrap $95\%$ intervals. Boxes
overstate the footprint of irregular objects, so bins extend above the
dataset's $5\%$ mask-area filter.}
\label{tab:size}
\begin{tabular}{@{}lrccc@{}}
\toprule
Object area & $n$ & Corrected & MINER & $\Delta$ (95\% CI) \\
\midrule
$<0.1\%$      & 1343 & 41.40 & 42.29 & $+0.89$ $[-0.45, +2.24]$ \\
$0.1$--$0.5\%$ & 2353 & 41.65 & 43.35 & $+1.70$ $[+0.68, +2.72]$ \\
$0.5$--$1\%$   & 1027 & 43.52 & 45.96 & $+2.43$ $[+0.78, +3.99]$ \\
$1$--$2\%$     &  887 & 48.93 & 50.62 & $+1.69$ $[+0.00, +3.49]$ \\
$2$--$5\%$     &  984 & 53.76 & 57.72 & $\mathbf{+3.96}$ $[+2.34, +5.79]$ \\
$5$--$10\%$    &  491 & 66.40 & 68.23 & $+1.83$ $[+0.00, +3.67]$ \\
$>10\%$        & 1146 & 75.65 & 78.36 & $+2.71$ $[+1.31, +4.19]$ \\
\bottomrule
\end{tabular}
\end{table}

\subsection{Stage Decomposition}
\label{app:decomposition}

Table~\ref{tab:decomposition} decomposes MINER into its two stages on
every backbone and split: each stage applied alone, and both together.
The pattern of Section~5.3 holds throughout: the stages are
near-additive, rescoring is the larger single stage everywhere, and the
crop stage's share grows on the ROCS splits, where the query names an
object the global embedding underweights.

\begin{table}[h]
\centering
\footnotesize
\setlength{\tabcolsep}{5pt}
\caption{Stage decomposition, text-to-image R@1. $+$CSLS and $+$crops
apply one stage alone; MINER applies both.}
\label{tab:decomposition}
\begin{tabular}{@{}llcccc@{}}
\toprule
Backbone & Method & ROCS-COCO & ROCS-Flickr & COCO\,5K & Flickr30K \\
\midrule
\multirow{4}{*}{CLIP\,L/14}
 & Baseline & 29.10 & 31.98 & 36.32 & 64.48 \\
 & $+$crops & 30.85 & 33.11 & 38.20 & 67.16 \\
 & $+$CSLS  & 34.93 & 37.14 & 42.67 & 72.30 \\
 & MINER    & \textbf{37.12} & \textbf{39.47} & \textbf{44.44} & \textbf{74.14} \\
\midrule
\multirow{4}{*}{SigLIP\,So/14}
 & Baseline & 45.41 & 46.18 & 54.24 & 82.94 \\
 & $+$crops & 47.05 & 48.91 & 55.23 & 83.84 \\
 & $+$CSLS  & 48.75 & 50.05 & 57.55 & 86.32 \\
 & MINER    & \textbf{50.59} & \textbf{52.15} & \textbf{58.22} & \textbf{86.90} \\
\midrule
\multirow{4}{*}{SigLIP\,2\,So/16}
 & Baseline & 47.08 & 48.00 & 56.55 & 83.72 \\
 & $+$crops & 49.11 & 50.27 & 57.35 & 84.50 \\
 & $+$CSLS  & 50.29 & 52.28 & 59.44 & 86.36 \\
 & MINER    & \textbf{52.36} & \textbf{53.84} & \textbf{60.24} & \textbf{87.08} \\
\bottomrule
\end{tabular}
\end{table}

\section{Hubness Mitigation}
\label{app:hubness_analysis}

\subsection{Inductive CSLS and Skewness Statistics}
\label{app:inductive}

Table~\ref{tab:inductive} broadens the held-out CSLS check of the main
paper to every backbone and split. The protocol scores the fused
similarities on a held-out half of the queries; the inductive variant
estimates the gallery-side CSLS term from the other half instead of the
scored queries. Inductive CSLS lands within $0.9$ R@1 of transductive
on all twelve combinations, within $0.3$ on nine, and above it on
seven, so the correction does not depend on scoring the test queries
jointly. For deployment this is the recommended configuration: estimate
the gallery-side term once from a historical query log, refresh it
periodically, and score each incoming query independently; the
query-side term needs only the query itself. These protocol studies run on cached embeddings with a
simplified crop extractor, so absolute values can differ from the main
table by up to $0.5$ R@1; each row is computed with one pipeline.

\begin{table}[H]
\centering
\footnotesize
\setlength{\tabcolsep}{4pt}
\caption{Fused-score R@1 on held-out queries under four CSLS protocols: no correction, transductive (both terms on the scored queries), inductive (gallery-side term from a disjoint query bank, query-side term over the full gallery), and the deployment variant of Section~5.4 of the main paper (bank plus query-side term over the top-$M$ shortlist), for $M{=}25$ and $M{=}100$.}
\label{tab:inductive}
\begin{tabular}{@{}llccccc@{}}
\toprule
Backbone & Split & None & Transductive & Inductive & $M{=}25$ & $M{=}100$ \\
\midrule
CLIP L/14 & ROCS-COCO & 30.22 & 35.81 & 36.52 & 36.47 & 36.52 \\
 & ROCS-Flickr & 33.12 & 37.88 & 37.81 & 37.55 & 37.81 \\
 & COCO 5K & 37.33 & 43.61 & 43.59 & 43.54 & 43.59 \\
 & Flickr30K & 67.80 & 73.16 & 74.00 & 74.00 & 74.00 \\
\midrule
SigLIP So/14 & ROCS-COCO & 47.16 & 49.17 & 49.44 & 49.47 & 49.44 \\
 & ROCS-Flickr & 47.62 & 50.86 & 50.86 & 50.74 & 50.86 \\
 & COCO 5K & 54.31 & 57.34 & 57.28 & 57.28 & 57.28 \\
 & Flickr30K & 83.52 & 85.20 & 85.68 & 85.68 & 85.68 \\
\midrule
SigLIP\,2\,So/16 & ROCS-COCO & 48.71 & 51.51 & 51.36 & 51.38 & 51.36 \\
 & ROCS-Flickr & 49.00 & 51.82 & 52.08 & 52.08 & 52.08 \\
 & COCO 5K & 57.48 & 59.18 & 59.48 & 59.49 & 59.48 \\
 & Flickr30K & 85.28 & 87.12 & 87.16 & 87.16 & 87.16 \\
\bottomrule
\end{tabular}
\end{table}

\subsection{QBNorm Temperature Sweep}
\label{app:qbnorm}

Table~\ref{tab:qbnorm-sweep} reports the complete QBNorm inverse-temperature
sweep summarized in the main paper: $\beta \in \{1, 2, 3, 5, 7, 10, 20\}$
on COCO\,5K for all three backbones, using the original formulation with
a self-normalizing query bank and global image embeddings. The response
to $\beta$ is smooth and consistent across backbones: a shallow optimum
at $\beta \in [2, 3]$, then a steady decline that steepens toward the
original default $\beta{=}20$. Even at its per-backbone optimum, QBNorm
recovers less of the available gain than two-sided CSLS at its single
untuned setting ($k{=}10$). The decline at large $\beta$ follows from
the $\exp(\beta\cos)$ weighting: on cosine-range similarities, large
$\beta$ concentrates the bank weights onto a handful of entries.

\begin{table}[h]
\centering
\footnotesize
\setlength{\tabcolsep}{4pt}
\caption{QBNorm inverse-temperature sweep, text-to-image R@1 on
COCO\,5K. Bold marks the best $\beta$ per backbone; CSLS is shown for
reference.}
\label{tab:qbnorm-sweep}
\begin{tabular}{@{}lccc@{}}
\toprule
 & CLIP\,L/14 & SigLIP\,So/14 & SigLIP\,2\,So/16 \\
\midrule
No correction        & 36.32 & 54.24 & 56.55 \\
\midrule
QBNorm $\beta{=}1$   & 37.96 & 54.65 & 56.93 \\
QBNorm $\beta{=}2$   & \textbf{38.20} & 54.91 & 57.15 \\
QBNorm $\beta{=}3$   & 38.04 & \textbf{55.07} & \textbf{57.29} \\
QBNorm $\beta{=}5$   & 35.77 & 54.54 & 57.09 \\
QBNorm $\beta{=}7$   & 32.40 & 53.37 & 56.25 \\
QBNorm $\beta{=}10$  & 26.43 & 51.11 & 54.47 \\
QBNorm $\beta{=}20$  & 13.08 & 42.87 & 47.43 \\
\midrule
CSLS $k{=}10$        & 42.67 & 57.55 & 59.44 \\
\bottomrule
\end{tabular}
\end{table}

\subsection{Sinkhorn Sensitivity}
\label{app:sinkhorn}

Table~\ref{tab:sinkhorn} sweeps Sinkhorn normalization over temperature
$\tau$ and iteration count on the fused scores, under the same held-out
protocol as the main-text comparison. The method is sharply peaked at
$\tau{=}0.02$: at a fixed iteration count, R@1 swings by $3$ to $8$
points across $\tau$, the spread widening with the iteration count. Iteration count matters only at small
$\tau$, where more iterations steadily hurt. The single setting that
beats CSLS on both splits is $\tau{=}0.02$ with one iteration, which is
barely Sinkhorn at all; at convergence the same $\tau$ wins on
ROCS-Flickr30K ($+0.9$) and loses on ROCS-COCO ($-0.9$). CSLS needs no
such tuning.

\begin{table}[h]
\centering
\footnotesize
\setlength{\tabcolsep}{5pt}
\caption{Sinkhorn R@1 by temperature and iteration count on fused
scores, held-out protocol. Reference: CSLS $k{=}10$ scores $51.51$
(ROCS-COCO) and $51.82$ (ROCS-Flickr30K); bold beats CSLS.}
\label{tab:sinkhorn}
\begin{tabular}{@{}lcccccc@{}}
\toprule
 & \multicolumn{6}{c}{Iterations} \\
\cmidrule(lr){2-7}
$\tau$ & 1 & 3 & 5 & 10 & 20 & 50 \\
\midrule
\multicolumn{7}{@{}l}{\emph{ROCS-COCO}} \\
0.005 & 50.46 & 49.73 & 48.66 & 46.82 & 44.39 & 42.42 \\
0.01  & 51.43 & 50.39 & 49.08 & 47.42 & 46.55 & 46.62 \\
0.02  & \textbf{51.75} & 51.21 & 50.92 & 50.58 & 50.66 & 50.66 \\
0.05  & 49.10 & 49.00 & 49.03 & 49.03 & 49.03 & 49.03 \\
0.1   & 47.72 & 47.69 & 47.69 & 47.69 & 47.69 & 47.69 \\
\midrule
\multicolumn{7}{@{}l}{\emph{ROCS-Flickr30K}} \\
0.005 & 50.74 & 50.22 & 49.41 & 47.58 & 45.65 & 44.68 \\
0.01  & 51.67 & 50.93 & 50.11 & 49.33 & 49.22 & 48.88 \\
0.02  & \textbf{53.49} & \textbf{52.90} & \textbf{52.79} & \textbf{52.71} & \textbf{52.68} & \textbf{52.68} \\
0.05  & 49.63 & 49.63 & 49.70 & 49.70 & 49.70 & 49.70 \\
0.1   & 48.18 & 48.22 & 48.22 & 48.22 & 48.22 & 48.22 \\
\bottomrule
\end{tabular}
\end{table}

\subsection{Hub Statistics}
\label{app:hubstats}

Table~\ref{tab:hubstats} quantifies the hubness the rescoring stage
corrects. Each gallery image's $k$-occurrence is the number of queries
that place it in their top $10$ under the fused score; a healthy gallery
has a low-skew occurrence distribution and few images that never appear.
Two-sided CSLS lowers the skewness and shrinks the unreachable set by
four times, while the single worst hub is reduced only modestly: the
correction rebalances the tail rather than removing individual hubs.

\begin{table}[h]
\centering
\footnotesize
\caption{Hub statistics on ROCS-COCO (SigLIP\,2\,So/16, $8{,}231$
queries over a $3{,}248$-image gallery, $k{=}10$). The $k$-occurrence of a
gallery image is the number of queries whose top-$10$ contains it; a
perfectly balanced gallery would give about $25$ per image. Lower skewness
means fewer hubs; the worst hub is the most-retrieved image; the last row
is the share of images that no query retrieves.}
\label{tab:hubstats}
\begin{tabular}{@{}lcc@{}}
\toprule
 & Fused score & \phantom{0}$+$ two-sided CSLS \\
\midrule
Skewness of $k$-occurrence (lower is better) & 2.35 & 1.80 \\
Worst hub ($k$-occurrence of the most-retrieved image) & 212 & 182 \\
Images in no query's top-$10$ & 1.6\% & 0.4\% \\
\bottomrule
\end{tabular}
\end{table}

\section{Inference Efficiency}
\label{app:efficiency_variants}

\subsection{Candidate Shortlist Top-$M$ Re-ranking}
\label{app:topm}

Table~\ref{tab:topm} reports the retrieve-then-rerank pipeline: candidate images are first retrieved by global embedding, and MINER is applied only to the top-$M$ shortlist. Small shortlists ($M{=}10$--$25$) recover the full-gallery retrieval gain across all four benchmark splits, showing that crop encoding can be delayed until inference without requiring a full regional index. At $M{=}500$, performance matches full-gallery MINER identically.

Crop embeddings are computed dynamically for shortlisted candidates, caching shared encodings across queries. Two-sided CSLS is then evaluated over the complete gallery matrix. At web scale, this architecture maintains a compact global index ($2.3$\,GB per million images vs.\ $14$\,GB for full crop storage) at $76$\,ms per query at $M{=}25$.

\begin{table}[h]
\centering
\footnotesize
\setlength{\tabcolsep}{5pt}
\caption{\textbf{Top-$M$ shortlist re-ranking} on SigLIP\,2 (R@1 and latency). Small $M$ ($10$--$25$) recovers full-gallery MINER performance.}
\label{tab:topm}
\begin{tabular}{@{}lccccc@{}}
\toprule
$M$ & ROCS-COCO & ROCS-Flickr & COCO\,5K & Flickr30K & ms / query \\
\midrule
base   & 50.29 & 52.28 & 59.44 & 86.36 & n/a \\
10     & 52.28 & 53.88 & \textbf{60.34} & 87.06 & 76 \\
25     & \textbf{52.50} & \textbf{53.95} & 60.32 & \textbf{87.10} & 76 \\
50     & 52.44 & 53.80 & 60.26 & 87.08 & 75 \\
100    & 52.35 & 53.82 & 60.24 & 87.08 & 75 \\
500    & 52.36 & 53.84 & 60.24 & 87.08 & 81 \\
\midrule
full   & 52.36 & 53.84 & 60.24 & 87.08 & n/a \\
\bottomrule
\end{tabular}
\end{table}

\subsection{Object-Level Attribution}
\label{app:attribution}

For every query MINER retrieves correctly, we check whether the
highest-scoring crop contains the annotated object (at least half of its
SAM3 box inside the crop). Chance is the fraction of the five crops that
contain the object, computed per query. On the ROCS-COCO queries where
MINER corrects the hubness-corrected baseline, the winning crop contains
the object in $68\%$ of cases against $29\%$ chance ($2.3\times$,
$n{=}343$; $65\%$ vs.\ $28\%$ on SigLIP So/14, $53\%$ vs.\ $29\%$ on
CLIP L/14); over all correctly retrieved queries, $52\%$ vs.\ $29\%$
($n{=}3{,}776$). Queries whose object lies outside every crop ($6\%$)
are excluded.

\paragraph{Object masking.} Attribution shows co-location, not
causation, so we also intervene on the image. On the $365$ ROCS-COCO
queries that MINER retrieves correctly and the hubness-corrected baseline
does not, we mask the named object's SAM3 box in the ground-truth image
(grey fill), re-encode the image with the same frozen encoder, and check
whether it is still ranked first; the control masks a random region of
the same size elsewhere in the image. Without masking, all $365$ stay at
rank~1.

\begin{table}[H]
\centering
\footnotesize
\caption{Object masking on the $365$ ROCS-COCO queries that MINER fixes:
share of queries whose correct image loses rank~1 (SigLIP\,2\,So/16; the
same test on the other backbones' corrected sets in the last two columns).}
\label{tab:masking}
\begin{tabular}{@{}lccc@{}}
\toprule
Masked region & SigLIP\,2 ($n{=}365$) & SigLIP So/14 ($n{=}374$) & CLIP L/14 ($n{=}388$) \\
\midrule
Named object            & 53\% & 50\% & 38\% \\
Random, same size       & 16\% & 12\% & 24\% \\
\bottomrule
\end{tabular}
\end{table}

The gap holds for the smallest objects (under $0.5\%$ of the image,
$n{=}155$: $31\%$ vs.\ $8\%$) and on the other backbones; CLIP's effect
is weaker on tiny objects, consistent with its $224$-px input. On a
random sample of $801$ queries that MINER gets right, most of which the
baseline also gets right, the rates fall to $22\%$ vs.\ $10\%$: the
object matters most in exactly the queries that MINER changes. The
remaining cases are consistent with the caption also describing the
scene around the object, which the crop still covers after masking.

\subsection{Crop-Fusion Blending Rules}
\label{app:fusion}

Table~\ref{tab:fusion} compares rules for aggregating the five crop
similarities before the blend of Equation~1, holding $\alpha{=}0.4$ and
CSLS $k{=}10$ fixed. Max pooling is within $0.1$ R@1 of the best rule
on both splits. Mean pooling costs $1.1$--$1.6$ R@1: typically a single
crop contains the rare object, and averaging dilutes its evidence with
four irrelevant crops. Softmax-weighted pooling approaches max as the
temperature drops and never exceeds it by more than $0.1$, so the paper
keeps the parameter-free max.

\begin{table}[h]
\centering
\footnotesize
\setlength{\tabcolsep}{5pt}
\caption{Crop-fusion rules on SigLIP\,2, text-to-image R@1.}
\label{tab:fusion}
\begin{tabular}{@{}lcc@{}}
\toprule
Fusion over crops & ROCS-COCO & ROCS-Flickr30K \\
\midrule
Max (default)        & 52.36 & 53.84 \\
Mean                 & 50.77 & 52.72 \\
Top-2 mean           & 51.83 & 53.69 \\
Softmax $T{=}0.05$   & 51.80 & \textbf{53.93} \\
Softmax $T{=}0.01$   & \textbf{52.46} & 53.88 \\
\bottomrule
\end{tabular}
\end{table}

\subsection{Gallery Scaling}
\label{app:gallery}

Holding the ROCS queries fixed, we grow the gallery with deduplicated
distractors from COCO\,5K, Flickr30K and the other ROCS split, to
$11{,}232$ images at the largest size ($3.5\times$ for the COCO queries,
$4.6\times$ for Flickr). Numbers are recomputed within this run, so the
smallest-gallery column differs slightly from the main tables.

\begin{table}[H]
\centering
\footnotesize
\setlength{\tabcolsep}{4pt}
\caption{Gallery scaling, R@1 as baseline / +CSLS / MINER. MINER stays ahead of the baseline on all $24$ cells; the region gain over the corrected baseline is $+1.5$ to $+2.4$ at every size.}
\label{tab:gallery}
\resizebox{\textwidth}{!}{%
\begin{tabular}{@{}llcccc@{}}
\toprule
Queries & Backbone & smallest & mid-1 & mid-2 & $11{,}232$ \\
\midrule
ROCS-COCO & CLIP L/14 & 29.10 / 34.93 / 37.02 & 24.14 / 28.27 / 30.08 & 23.39 / 27.40 / 29.21 & 21.78 / 25.65 / 27.38 \\
 & SigLIP So/14 & 45.41 / 48.75 / 50.26 & 40.86 / 42.29 / 44.03 & 40.41 / 41.50 / 43.28 & 38.74 / 39.61 / 41.32 \\
 & SigLIP\,2\,So/16 & 47.08 / 50.29 / 52.02 & 42.47 / 43.14 / 45.17 & 42.00 / 42.28 / 44.27 & 40.54 / 40.51 / 42.12 \\
\midrule
ROCS-Flickr & CLIP L/14 & 31.98 / 37.14 / 39.32 & 29.13 / 33.67 / 35.82 & 25.79 / 29.02 / 31.07 & 24.48 / 27.03 / 29.13 \\
 & SigLIP So/14 & 46.18 / 50.05 / 52.39 & 44.56 / 47.07 / 49.49 & 40.90 / 42.54 / 44.75 & 38.20 / 40.06 / 41.92 \\
 & SigLIP\,2\,So/16 & 48.00 / 52.28 / 53.84 & 46.07 / 48.75 / 50.79 & 42.28 / 43.43 / 45.84 & 39.54 / 41.05 / 43.06 \\
\bottomrule
\end{tabular}}
\end{table}

Gallery sizes are $3{,}248$ / $7{,}862$ / $8{,}862$ / $11{,}232$ for the
COCO queries and $2{,}442$ / $3{,}370$ / $8{,}370$ / $11{,}232$ for the
Flickr queries. The hubness share shrinks as the gallery diversifies
(SigLIP\,2, COCO queries: $+3.21$, $+0.67$, $+0.28$, $-0.03$), while the
region gain holds.

\subsection{Budget Accounting}
\label{app:budget}

Table~\ref{tab:budget} lists stored embeddings per image and encoder
passes for every region-aware alternative in the paper, all on the same
frozen encoder and all hubness-corrected. Rerankers that need a
cross-encoder or a detector fall outside this setting.

\begin{table}[H]
\centering
\footnotesize
\caption{Storage and compute accounting, R@1 on ROCS-COCO / ROCS-Flickr30K (SigLIP\,2\,So/16).}
\label{tab:budget}
\begin{tabular}{@{}lccc@{}}
\toprule
Method & Stored emb. / image & Encoder passes & R@1 \\
\midrule
Global only & 1 & 1 & 50.29 / 52.28 \\
5 random crops & 6 & 6 & 51.06 / 52.84 \\
5 attention crops & 6 & 6 & 51.40 / 53.13 \\
$3{\times}3$ grid & 10 & 10 & 51.96 / 53.69 \\
Dense patch tokens & 576 & 1 & 29.1 / n/a \\
MINER, fixed 5 crops & 6 & 6 & 52.36 / 53.84 \\
MINER, top-25 rerank & 1 & 1 offline + 125 online & 52.50 / 53.95 \\
\bottomrule
\end{tabular}
\end{table}

%% file: sec/appendix_audit.tex
\section{ROCS Human Audit}
\label{app:audit}

\begin{figure}[H]
\centering
\includegraphics[width=0.8\textwidth]{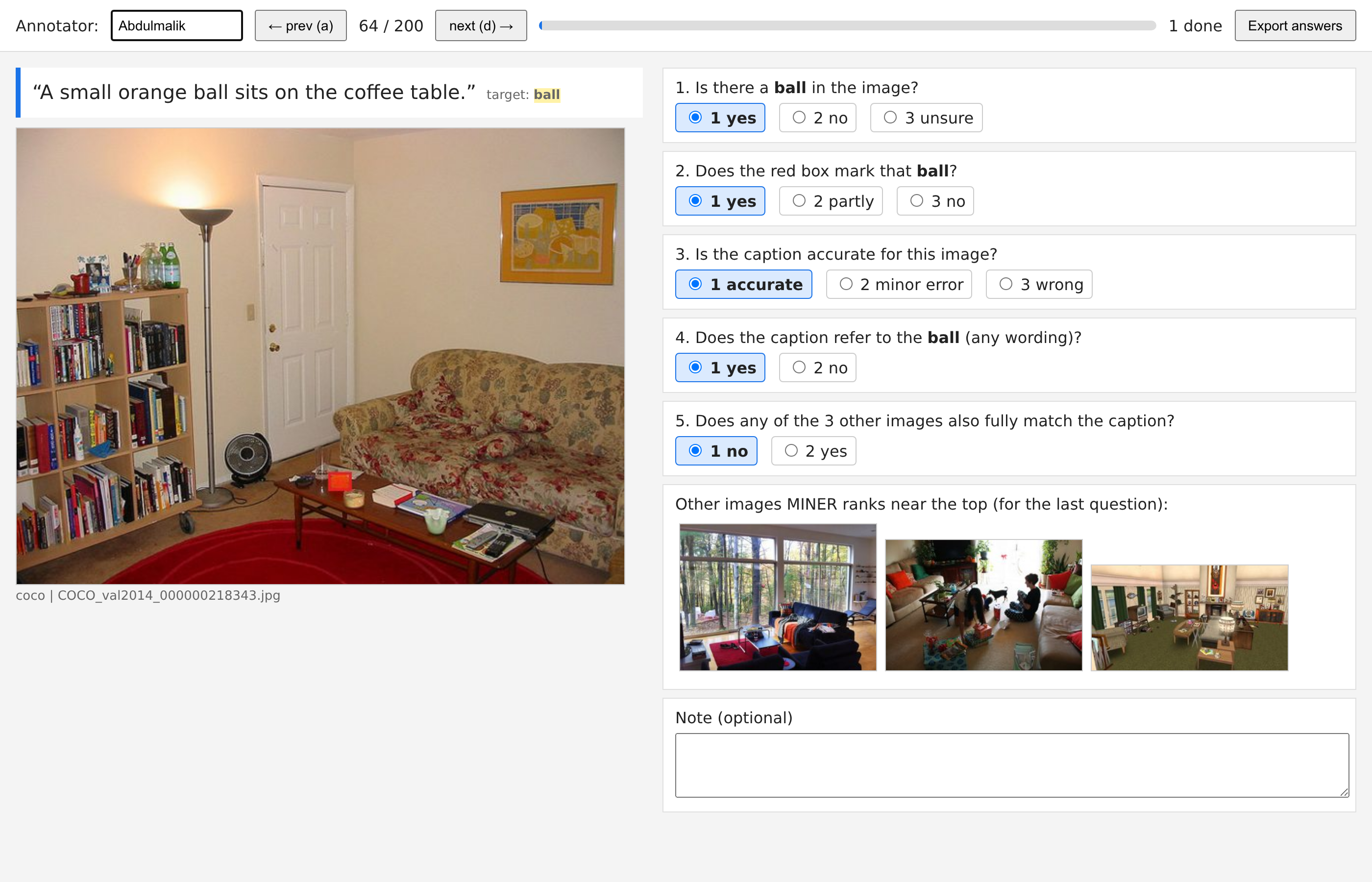}
\caption{The ROCS audit tool. Left: the caption with the target class
highlighted and the image with the target's SAM3 box. Right: the five
questions, the three other images MINER ranks highest, and a free-text
note.}
\label{fig:audit_tool}
\end{figure}

An author audited $200$ random ROCS queries ($100$ per split, seed $0$)
with the tool in Figure~\ref{fig:audit_tool}, answering for each whether
the named object is present, whether the SAM3 box marks it (ROCS-COCO),
whether the caption is accurate and refers to the object, and whether the
caption also fully fits one of the three highest-ranked other gallery
images. Table~\ref{tab:audit} reports the three headline rates.
$12\%$ of captions contain a wrong claim; $37\%$ of captions also
fully fit another top-ranked image, and on those queries R@1 is $19\%$
for the global baseline and $16\%$ for MINER against $59\%$ and $65\%$
on the rest.

\begin{table}[H]
\centering
\footnotesize
\caption{ROCS audit on $200$ random queries, $100$ per split.}
\label{tab:audit}
\begin{tabular}{@{}llc@{}}
\toprule
Aspect & Criterion & Agreement \\
\midrule
Object existence     & The queried object is present in the image                & $82\%$ \\
Caption faithfulness & The caption is correct, allowing a minor detail           & $88\%$ \\
Target relevance     & The caption refers to the queried object                  & $78\%$ \\
\bottomrule
\end{tabular}
\end{table}

%% file: sec/appendix_figures.tex
\section{Region Candidate Generators}
\label{app:cropmethods}

Figure~\ref{fig:cropmethods} illustrates the region-candidate
generators ablated in the main paper (Table~1): the fixed five-crop layout
(center and four corners at $60\%$ scale), a $3{\times}3$ uniform grid,
$N$ random crops, and $N$ attention-placed crops, together with a
saliency-guided variant that replaces the center crop by one crop placed
on a saliency peak. The top row shows candidate placements on the source
image; the columns below show the resulting crops fed to the frozen
encoder. All $60\%$-scale generators land within $1.3$ R@1 of one
another, consistent with spatial coverage rather than precise
localisation driving the gains.

\section{Saliency-source comparison}
\label{app:saliency}
MaskCLIP, DINOv3, and CLIP-Surgery produce visibly different attention
maps (shown in the main paper), yet swapping the source that places the
guided crop in the saliency-guided variant (center crop replaced by one saliency-placed crop) changes R@1 by at most $0.42$ in any
cell (Table~\ref{tab:saliency}).

\begin{table}[htbp]
\centering
\footnotesize
\setlength{\tabcolsep}{5pt}
\caption{R@1 when swapping the saliency source that places the guided
crop; fixed five-crop reference in the last row.}
\label{tab:saliency}
\begin{tabular}{@{}lcccc@{}}
\toprule
Saliency source & ROCS-COCO & ROCS-Flickr30K & COCO\,5K & Flickr30K \\
\midrule
Own attention & 52.18 & 53.63 & 59.96 & 86.74 \\
MaskCLIP      & 52.02 & 53.67 & \textbf{60.06} & 86.84 \\
DINOv3       & \textbf{52.44} & 53.54 & \textbf{60.06} & \textbf{86.92} \\
CLIP-Surgery & 52.23 & \textbf{53.69} & 59.96 & 86.84 \\
\midrule
Fixed 5-crop (no saliency) & 52.36 & 53.84 & 60.24 & 87.08 \\
\bottomrule
\end{tabular}
\end{table}

%% file: sec/appendix_prompts.tex
\section{Dataset Prompts and Vocabulary}
\label{app:dataset_prompts}

\subsection{Qwen3-VL Captioning Prompts}
\label{app:prompts}

This section documents the exact prompts used to generate MS COCO--style
captions during the curation of the ROCS dataset. The placeholder
\texttt{\{rare\_class\}} is substituted at runtime with the highlighted
object category for each image.

\subsubsection{System Prompt}

\begin{quote}
\ttfamily\small\raggedright
You write English image captions in the style of the MS COCO dataset: exactly one clear sentence ending with a period; neutral and factual.\\[0.5em]
Target length about 10--18 words, not longer. Prefer a single short clause (e.g. ``A woman holding skis on a snowy slope.''), not chained detail or lists.\\[0.5em]
The user cares about one highlighted object/category. Mention it naturally; you may paraphrase awkward labels (``fork spoon'', ``bowl plate'') into normal wording.\\[0.5em]
Avoid extra adjectives, double clauses, commas lists, clich\'es (``gleaming blade'', ``slightly worn''), theatrical tone, or multiple sentences.\\[0.5em]
Do not start with meta phrases (``This image'', ``The caption''). Output only the caption sentence.
\end{quote}

\subsubsection{User Prompt}

\begin{quote}
\ttfamily\small\raggedright
Write one MS COCO--style caption for this photo: one short sentence only, natural and neutral, that clearly includes what the label \texttt{\{rare\_class\}} refers to (rephrase if needed). Roughly 8--16 words, not more.\\[0.5em]
Caption:
\end{quote}

\subsection{SAM3 Detection Prompts List}
\label{app:vocab}

We prompt SAM~3 with the canonical 80 MS COCO class names, issued
verbatim as noun-phrase queries, with six labels rephrased to disambiguate them in an
open-vocabulary setting where the detector grounds free-form text rather
than a closed label index. Specifically, we rename
\texttt{sports ball} $\rightarrow$ \texttt{ball},
\texttt{orange} $\rightarrow$ \texttt{orange fruit},
\texttt{tv} $\rightarrow$ \texttt{television},
\texttt{mouse} $\rightarrow$ \texttt{computer mouse},
\texttt{remote} $\rightarrow$ \texttt{remote control}, and
\texttt{keyboard} $\rightarrow$ \texttt{computer keyboard}. The
remaining $74$ prompts are the canonical class names unchanged.

%% file: sec/appendix_cropfig.tex
\begin{figure}[h]
\centering
\includegraphics[width=\textwidth]{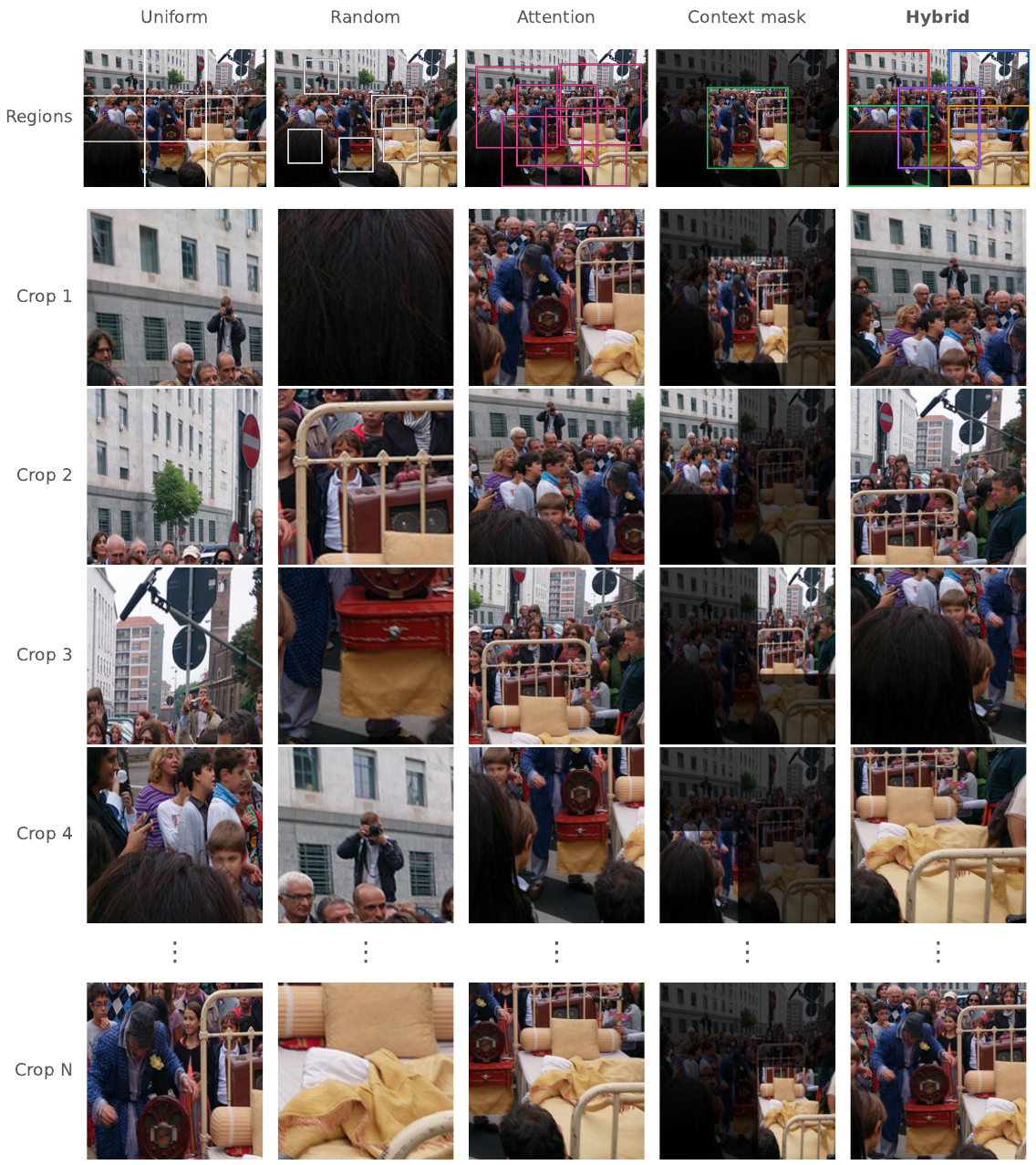}
\caption{Region candidate generators. Top row: candidate placements;
columns below: the resulting crops.}
\label{fig:cropmethods}
\end{figure}